\documentclass{article}

\usepackage{iclr2027_conference,times}

\usepackage{microtype}
\usepackage{graphicx}
\usepackage{subcaption}
\usepackage{booktabs}
\usepackage{seqsplit}

\usepackage{amsmath}
\usepackage{amssymb}
\usepackage{mathtools}
\usepackage{amsthm}

\usepackage{multirow}
\usepackage{tcolorbox}
\usepackage[dvipsnames,table,xcdraw]{xcolor}
\usepackage[normalem]{ulem}
\useunder{\uline}{\ul}{}

\usepackage{capt-of} 

\usepackage[textsize=tiny]{todonotes}

\usepackage{hyperref}
\usepackage{url}

\hypersetup{
    hypertex=true,
    colorlinks=true,
    linkcolor=[RGB]{105,33,106},
    anchorcolor=[RGB]{105,33,106},
    citecolor=[RGB]{105,33,106},
    urlcolor=[RGB]{0,76,129}
}

\usepackage[capitalize,noabbrev]{cleveref}

\theoremstyle{plain}

\theoremstyle{definition}

\theoremstyle{remark}

\newcommand{\method}{NAMOH}

\title{
Scaling Parameter and Context in Attention: Native Sparse Attention from Mixture-of-Head
}

\author{%
    Zizhuo Fu\\
    Peking University\\
    \And
    Runsheng Wang \\
    Peking University \\
    \And
    Meng Li \\
    Peking University \\
}

\iclrfinalcopy

\begin{document}

\maketitle
\lhead{Native Sparse Attention from Mixture-of-Head}


\begin{abstract}
Scaling attention parameters can improve language model quality, but retaining full token histories makes additional heads costly at long contexts.
Furthermore, since attention retrieves and combines contextual information, parameter scaling should also support longer contexts.
We therefore ask \textit{whether attention parameter scaling can directly enable efficient and effective context scaling.}
We introduce \method{}, an architecture-native sparse attention mechanism that activates $K$ of $H$ heads per token.
Each head retains only its assigned tokens and performs causal attention within this subsequence.
Head selection thus jointly determines active parameters and available context without scanning the full history.
Under balanced assignments, increasing $H$ at fixed $K$ shortens head histories and reduces per-token key-value (KV) access without increasing total KV storage.
We further support head-relative rotary position embeddings to shorten positional spans within routed subsequences, aiming to mitigate position-induced attention noise.
Experiments show that \method{} can outperform fully activated models with the same total parameters, while enabling more efficient long-context inference than smaller dense models with matched active parameter counts.
It remains compatible with GQA and existing sparse attention mechanisms.
We hope this work offers a new path for scaling attention, with parameter scaling directly enabling context scaling.
\end{abstract}
\section{Introduction}
\label{sec:introduction}

\begin{figure}[t]
  \centering
  \vspace{-0.4cm}
  \includegraphics[width=\linewidth]{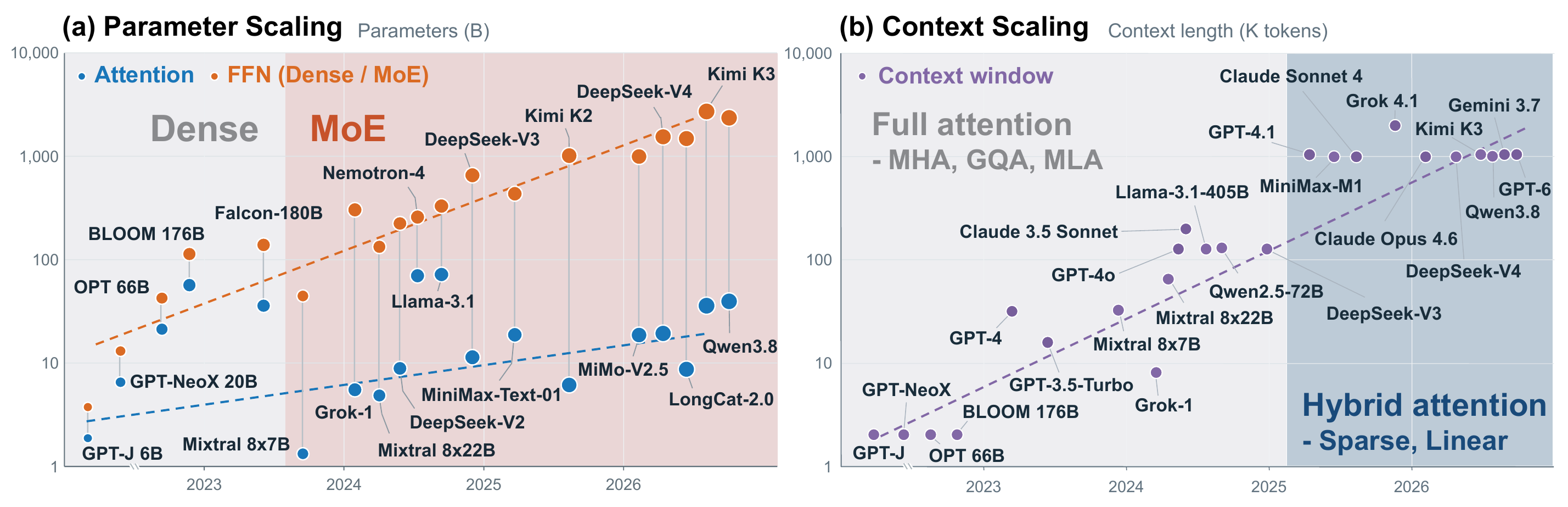}
  \vspace{-0.6cm}
  \caption{
    \textbf{Parameter and context scaling in frontier language models.}
    (a) Attention and FFN parameters in frontier models.
    Since the rise of MoE, parameter growth has been dominated by FFN experts, with comparatively little growth in attention.
    (b) Maximum context windows in frontier models.
    Sparse and linear-time mechanisms support further context expansion.
  }
  \vspace{-0.4cm}
  \label{fig:intro-scaling}
\end{figure}

Scaling attention parameters through expert routing can improve the performance of large language models (LLMs)~\citep{Zhang2022MixtureOA,Yang2025UMoEUA}.
Methods such as SwitchHead and MoH further demonstrate gains in model quality and efficiency by selectively activating attention heads or projections~\citep{Csords2023SwitchHeadAT,Jin2024MoHMA}.
Yet most parameter growth in frontier models has been concentrated in feed-forward networks (FFNs), particularly with the rise of mixture-of-experts (MoE) architectures~\citep{Jiang2024MixtralOE,Dubey2024TheL3,Adler2024Nemotron43T,mimov25,Yang2024Qwen25TR}.
As shown in Figure~\ref{fig:intro-scaling}(a), attention parameters have scaled much more slowly.

A key challenge is to increase attention parameters without proportionally increasing the cost of storing and processing context.
When each head retains its own full token history, adding heads increases key-value (KV) storage, while activating more heads increases full-attention computation.
This raises activation memory and compute costs during training and prefill, as well as KV cache storage and memory traffic during decoding~\citep{Dao2023FlashAttention2FA,Tang2024QuestQS}.
Yet efficiency is only part of the goal.
Mechanistic studies show that FFNs can store knowledge acquired during training~\citep{Geva2020TransformerFL}, while attention heads play a central role in retrieving and combining information from the current context~\citep{Olsson2022IncontextLA,Guo2024ActiveDormantAH}.
This role makes long-context processing a natural objective for attention parameter scaling.

Despite this connection, work on context scaling has primarily focused on computational efficiency rather than attention parameter scaling.
The expanding context windows in Figure~\ref{fig:intro-scaling}(b) have been supported by sparse and linear-time mechanisms, often combined with full attention~\citep{Yuan2025NativeSA,Yang2024GatedDN,DeepSeekAI2026DeepSeekV4TH,Bai2026KimiKO}.
Linear-time recurrent architectures compress history into fixed-size states, which can lose information needed for precise recall~\citep{Jelassi2024RepeatAM,Cabannes2026SparseDM}.
Selection-based sparse attention preserves explicit token memories but introduces token or block selection overhead~\citep{Lu2025MoBAMO,Yuan2025NativeSA}.
For indexers that scan the full prefix, this overhead grows with context length~\citep{Xu2026HISAEH}.
Computational savings alone also leave positional limitations unresolved.
As~\citet{Du2026RoPEDN} show, long contexts can yield indistinguishable attention scores for different positions or tokens and disrupt token relevance rankings.
Related geometric analysis links long RoPE spans to spurious query-key alignment, which can introduce attention noise by assigning weight to irrelevant tokens~\citep{Wertheimer2026FrayedRA}.

Together, these challenges motivate a joint view of attention parameter and context scaling.
The goal is not only to add attention parameters at a manageable cost, but also to use those parameters to support longer contexts.
We therefore ask: \textbf{\emph{Can scaling attention parameters directly enable efficient and effective context scaling?}}

We introduce \method{}, an architecture-native sparse attention mechanism that couples attention parameter and context scaling.
Each attention head acts as an expert, and a learned router activates the $K$ highest-scoring heads out of $H$ total heads for each token.
Routing determines both the active head parameters and the token subsequence processed by each head.
Each head performs causal attention only within its routed subsequence, and the layer combines the projected head outputs using routing weights.
Attention sparsity therefore arises directly from the routing structure, rather than masking a dense attention result.

We use load balancing to discourage head collapse~\citep{Jin2024MoHMA,Fu2026AttentionSF} and distribute tokens approximately evenly across heads.
With $H$ total heads and $K$ active heads per token, a context of $L$ tokens requires $LK$ key-value (KV) entries in total, with approximately $LK/H$ entries per head.
Each query token therefore accesses approximately $LK^2/H$ KV entries across its $K$ active heads.
We define the \emph{KV activation ratio} as this per-token KV access relative to a full-attention baseline.
For baselines with the same model width and head dimension, this ratio is approximately $K/H$ relative to $K$-head full attention matched in active attention parameters, and $(K/H)^2$ relative to $H$-head full attention matched in total attention parameters.

For example, with $H=32$ and $K=8$, total KV storage is $1/4$ that of 32-head full attention and equal to that of 8-head full attention.
Under balanced routing, each head retains approximately $L/4$ entries, so eight active heads access approximately $2L$ entries per token.
This yields KV activation ratios of $1/16$ and $1/4$ relative to the two baselines, respectively.
Parameter scaling thus allows more heads to specialize in different contextual patterns, with each head processing its routed subsequence.

We further support \emph{head-relative RoPE}, which encodes each token by its position within a head's routed subsequence rather than its global position.
By shortening the encoded span to approximately $LK/H$ under balanced routing while preserving token order, this design aims to mitigate position-induced attention noise.
For efficient training and prefill, we pack head-specific subsequences for variable-length FlashAttention~\citep{Dao2023FlashAttention2FA}.
During decoding, we execute only active head queries and read or update only their corresponding KV caches.

Our experiments show that \method{} can achieve higher accuracy than fully activated models with the same total parameter count.
Its reduced KV access also enables more efficient long-context inference than smaller dense models matched in active parameter count.
Moreover, \method{} is compatible with GQA and existing sparse attention mechanisms, which can further select tokens or blocks within each head's routed subsequence.
Together, these properties offer a complementary path to context scaling through attention parameter scaling.
\section{Background}
\label{sec:background}

\subsection{Head Sparsity: Mixture-of-Head}
\label{sec:background-moh}

Mixture-of-experts (MoE) layers scale feed-forward capacity through conditional activation without proportional growth in per-token computation~\citep{Shazeer2017OutrageouslyLN,Fedus2021SwitchTS}.
The same principle applies to attention parameters.
\citet{Peng2020AMO} mix overlapping groups of heads rather than route each token to individual heads.
Mixture of Attention Heads (MoA) routes query and output projections with shared keys and values~\citep{Zhang2022MixtureOA}, while SwitchHead routes value and output projections~\citep{Csords2023SwitchHeadAT}.
Grouped Query Experts (GQE) routes query heads within grouped-query attention~\citep{Tripathi2026GroupedQE}.

Mixture-of-Head attention (MoH) selects heads per token and weights their projected outputs~\citep{Jin2024MoHMA}.
However, its heads retain full key-value (KV) histories even for tokens whose head outputs are inactive.
With independent, fixed-width KV heads, cache storage grows with head count and context length, while each active query still processes the full prefix.
Mixture of Sparse Attention (MoSA) introduces sequence sparsity through expert-choice routing, with each head selecting a fixed quota of top-scoring tokens from the full sequence~\citep{Piekos2025MixtureOS}.
This balances head loads, but future tokens can change earlier selections despite causal attention masking, so incremental autoregressive decoding requires routing changes.

\subsection{Context Sparsity: Sparse Attention}
\label{sec:background-sa}

Selection-based sparse attention moves relevance filtering before the main softmax attention.
Quest ranks KV pages using query-dependent scores from key metadata~\citep{Tang2024QuestQS}.
Native Sparse Attention (NSA) reuses compressed attention scores to select blocks~\citep{Yuan2025NativeSA}, while Mixture of Block Attention (MoBA) selects blocks using query affinities to pooled keys~\citep{Lu2025MoBAMO}.
DeepSeek-V4 combines KV compression with a lightweight indexer in its compressed sparse attention branch~\citep{DeepSeekAI2026DeepSeekV4TH}.
When indexers scan the prefix, their overhead grows with context length~\citep{Xu2026HISAEH}.
Realizing speedups also requires specialized kernels and cache layouts for irregular KV access~\citep{DeepSeekAI2026DeepSeekV4TH}.
NSA also reports higher average performance than its full-attention baseline on general and long-context benchmarks, supporting sparsity as a modeling choice~\citep{Yuan2025NativeSA}.

\method{} provides a complementary source of sparsity through token-to-head assignments, without ranking the full history per query.
Within routed subsequences, softmax attention remains compatible with dynamic token selection~\citep{Zhang2023H2OHO,Tang2024QuestQS} and static patterns such as sliding windows with retained sink tokens~\citep{Xiao2023EfficientSL}.

Token selection does not necessarily shorten positional spans when original indices are retained.
For rotary position embeddings (RoPE), \citet{Wertheimer2026FrayedRA} link length extrapolation to disrupted query-key separation and spurious attention to irrelevant tokens.
\citet{Du2026RoPEDN} further identify indistinguishable scores across positions or tokens and reversed relevance rankings.
Sparsity alone therefore does not ensure reliable scores among retained tokens.
These findings motivate reducing the effective range of relative positions as a potential way to mitigate position-induced attention noise.
\section{\method{}: Native Sparse Attention from Mixture-of-Head}
\label{sec:namoh}

\begin{figure}[t]
  \centering
  \vspace{-0.4cm}
  \includegraphics[width=\linewidth]{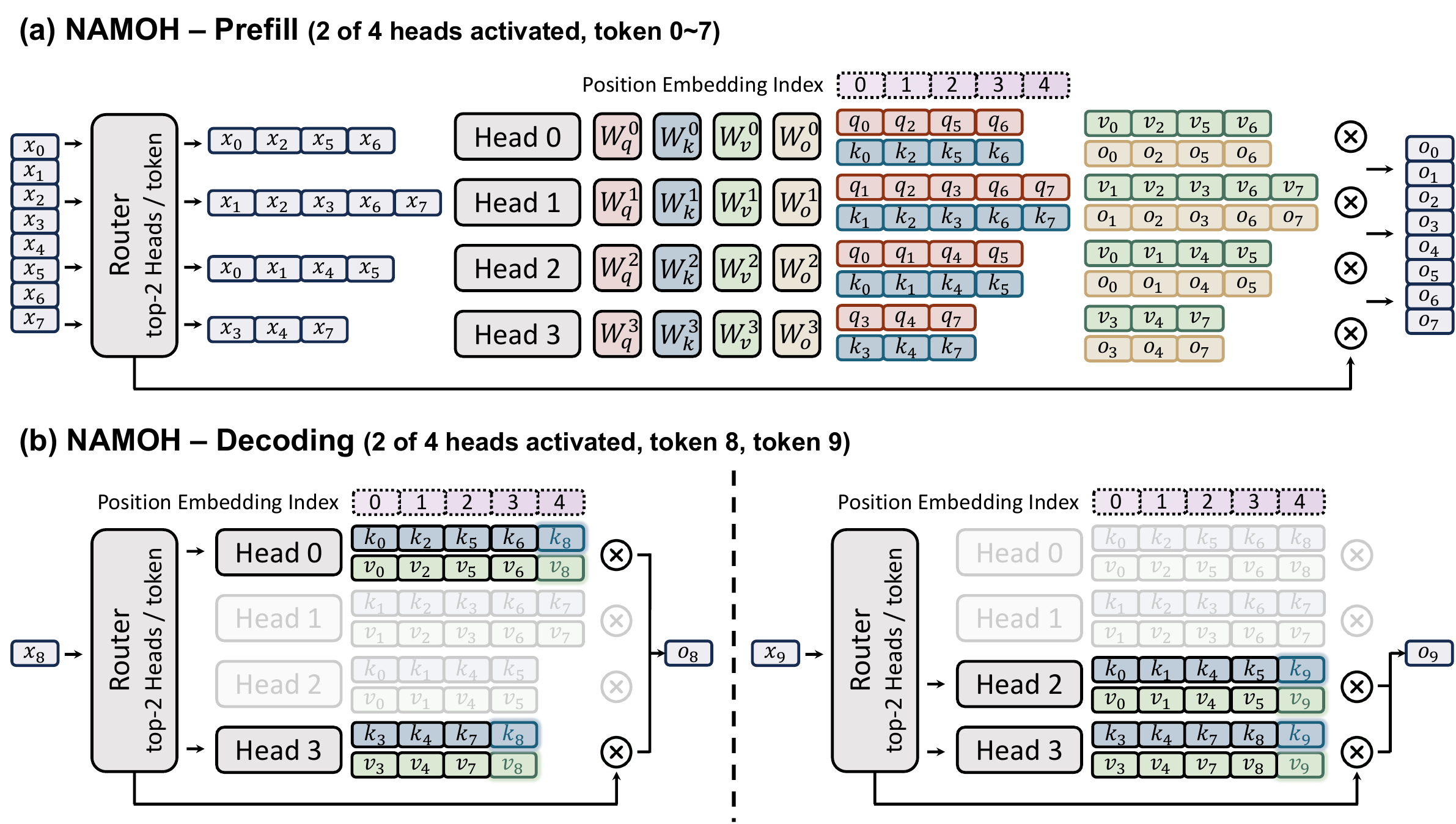}
  \vspace{-0.3cm}
  \caption{
    \textbf{\method{} with two active heads out of four.}
    (a) Prefill routes tokens $0$ to $7$ before QKV projection. Each head attends within its ordered subsequence, and projected outputs are combined with routing weights.
    (b) During decoding, tokens $8$ and $9$ access and extend only their selected heads' KV caches.
    Position indices follow each head's local token order.
  }
  \label{fig:method-namoh}
\end{figure}

\subsection{Routed Causal Attention}
\label{sec:namoh-routing}

For a length-$T$ sequence with token representations $x_t\in\mathbb{R}^{1\times d_{\mathrm{model}}}$ and model width $d_{\mathrm{model}}$, standard multi-head attention with $H$ heads~\citep{Vaswani2017AttentionIA} can be written as
\begin{equation}
    o_t^{\mathrm{full}}
    =\sum_{i=0}^{H-1}h_{t,i}^{\mathrm{full}}W_O^{(i)},
    \label{eq:namoh-dense}
\end{equation}
where $h_{t,i}^{\mathrm{full}}\in\mathbb{R}^{1\times d_{\mathrm{head}}}$ is head $i$'s causal attention output with head width $d_{\mathrm{head}}$, and $W_O^{(i)}\in\mathbb{R}^{d_{\mathrm{head}}\times d_{\mathrm{model}}}$ is its slice of the output projection.

\noindent\textbf{Token-to-head routing.}
Following the expert view of attention heads~\citep{Jin2024MoHMA}, \method{} activates $K$ of $H$ heads per token, where $1\leq K\leq H$.
A learned router computes head affinities, selections, and gates as
\begin{equation}
    a_t=\operatorname{softmax}(x_tW_r),\quad
    \mathcal{S}_K(t)=\operatorname{TopK}_{i}(a_{t,i}),\quad
    m_{t,i}=\mathbf{1}\{i\in\mathcal{S}_K(t)\},\quad
    g_{t,i}=m_{t,i}a_{t,i}.
    \label{eq:namoh-gates}
\end{equation}
Here, $W_r\in\mathbb{R}^{d_{\mathrm{model}}\times H}$ is the router matrix, and $a_t\in\mathbb{R}^{1\times H}$ contains affinities normalized by softmax across heads.
$\operatorname{TopK}$ returns the indices of the $K$ largest affinities, and $\mathbf{1}\{\cdot\}$ is the indicator function.
The selection mask $m_{t,i}$ sets inactive gates to zero.

\noindent\textbf{Sparse projection and attention.}
Routing determines both which parameters a token uses and which head histories it enters.
For head $i$, define the ordered index set $\mathcal{I}_i=\{t:m_{t,i}=1\}$ and its length $n_i=|\mathcal{I}_i|$.
Tokens are dispatched to these subsequences before projection.
Only an active token-head pair computes
\begin{equation}
    q_{t,i}=x_tW_Q^{(i)},\qquad
    k_{t,i}=x_tW_K^{(i)},\qquad
    v_{t,i}=x_tW_V^{(i)},\qquad t\in\mathcal{I}_i.
    \label{eq:namoh-qkv}
\end{equation}
Each head has independent query, key, and value matrices $W_Q^{(i)},W_K^{(i)},W_V^{(i)}\in\mathbb{R}^{d_{\mathrm{model}}\times d_{\mathrm{head}}}$.
The resulting queries, keys, and values lie in $\mathbb{R}^{1\times d_{\mathrm{head}}}$.
Inactive token-head pairs generate no QKV vectors and occupy no KV cache entries.

Let $\widetilde q_{t,i}$ and $\widetilde k_{t,i}$ denote queries and keys after positional encoding, described in Section~\ref{sec:namoh-properties}.
For $i\in\mathcal{S}_K(t)$, attention and output aggregation are
\begin{equation}
    h_{t,i}
    =\operatorname{Attn}\!\left(
        \widetilde q_{t,i},
        [\widetilde k_{s,i}],
        [v_{s,i}]
    \right), s\in\mathcal{I}_i,s\leq t, \qquad
    o_t=\sum_{i\in\mathcal{S}_K(t)}g_{t,i}h_{t,i}W_O^{(i)},
    \label{eq:namoh-attention}
\end{equation}
where $\operatorname{Attn}$ denotes standard scaled dot-product attention, $h_{t,i}\in\mathbb{R}^{1\times d_{\mathrm{head}}}$, $W_O^{(i)}\in\mathbb{R}^{d_{\mathrm{head}}\times d_{\mathrm{model}}}$, and $o_t\in\mathbb{R}^{1\times d_{\mathrm{model}}}$.
Each query therefore attends only to earlier tokens and itself within the same routed head.
Sparsity is part of the computation, not a mask applied after dense attention.

Figure~\ref{fig:method-namoh}(a) illustrates dispatch and aggregation with $H=4$ and $K=2$.
During decoding in Figure~\ref{fig:method-namoh}(b), token $8$ selects heads $0$ and $3$, while token $9$ selects heads $2$ and $3$.
Each token appends to and attends within only its selected caches.
Because routing uses the current token representation, prefill and incremental decoding implement the same causal computation.

\subsection{Load Balancing and Scaling Properties}
\label{sec:namoh-properties}

\noindent\textbf{Load balancing.}
Balancing encourages all heads to receive training signals and discourages head collapse~\citep{Fu2026AttentionSF}.
It also supports context sparsity: if routing concentrates on a fixed subset of heads, their histories can become dense despite sparse head activation.
For a training batch $\mathcal{B}$ containing $N$ tokens, define the assignment fraction $f_i$ and average head affinity $p_i$ as
\begin{equation}
    f_i=\frac{1}{NK}\sum_{t\in\mathcal{B}}m_{t,i},\qquad
    p_i=\frac{1}{N}\sum_{t\in\mathcal{B}}a_{t,i},
    \label{eq:namoh-balancing-statistics}
\end{equation}
with batch indices suppressed.
Both distributions sum to one, with uniform targets $f_i=p_i=1/H$.

We consider three alternatives from MoE: CV-based importance regularization~\citep{Shazeer2017OutrageouslyLN}, Switch-style ($fp$) balancing~\citep{Fedus2021SwitchTS}, and auxiliary-loss-free balancing~\citep{Wang2024AuxiliaryLossFreeLB}.
For the two auxiliary-loss methods, the training objective is
$\mathcal{L}=\mathcal{L}_{\mathrm{LM}}+\lambda_{\mathrm{bal}}\mathcal{L}_{\mathrm{bal}}$, where $\mathcal{L}_{\mathrm{LM}}$ is the language-modeling loss, $\mathcal{L}_{\mathrm{bal}}$ is the chosen balancing loss summed across routed attention layers, and $\lambda_{\mathrm{bal}}>0$ controls its strength.
Loss-free balancing instead adjusts head-specific routing biases with an update rate $\eta>0$, without adding a balancing loss.
Appendix~\ref{app:load-balancing} details all three strategies.

\noindent\textbf{Routing as context selection.}
Each head maintains its own routed history, so selecting heads also selects the contexts available to a query.
The head router thus serves as a learned indexer over histories without rescoring historical keys or blocks.
Unlike a fixed KV budget, the available history grows with sequence length, reaching approximately $TK/H$ entries in each head under balanced routing.
This selection acts at the head level and remains compatible with further token or block sparsity within each subsequence~\citep{Tang2024QuestQS,Yuan2025NativeSA,Lu2025MoBAMO}.

\noindent\textbf{Parameter and context costs in attention.}
We compare one attention layer with $H$ query heads at fixed model and head widths, abbreviated as $d_m$ and $d_h$.
Let $M_{\mathrm{KV}}$ count stored KV pairs after a prefix of $T$ tokens, and let $A_{\mathrm{KV}}(T)$ count distinct historical pairs used to decode the next token.
Each pair contains $2d_h$ scalars; shared pairs are counted once.
GQA shares one KV head among $g$ query heads, reducing both counts to $HT/g$~\citep{Ainslie2023GQATG}.
Block-selected sparse attention (SA) retains the full cache but accesses approximately an $\alpha\in(0,1]$ fraction of each history~\citep{Tang2024QuestQS,Lu2025MoBAMO}.
Thus, SA stores $HT$ pairs and activates approximately $\alpha HT$.
MoH instead retains all head histories and activates $KT$ pairs through its $K$ selected heads~\citep{Jin2024MoHMA}.

For \method{}, let $n_i$ denote head $i$'s prefix length and $\mathcal{S}_K(T)$ the next token's $K$ selected heads.
Routing determines both cache insertion and access:
\begin{equation}
    M_{\mathrm{KV}}=TK,\qquad
    A_{\mathrm{KV}}(T)
    =\sum_{i\in\mathcal{S}_K(T)}n_i
    \approx\frac{TK^2}{H}.
    \label{eq:namoh-kv-cost}
\end{equation}
Storage is exact, while activation assumes approximately balanced histories.
Relative to $H$-head MHA, these costs are reduced to $K/H$ and approximately $(K/H)^2$, respectively.
Relative to $K$-head MHA, which matches active projection parameters, storage is unchanged and activation is reduced to approximately $K/H$.

Table~\ref{tab:attention-complexity} compares attention projection weights and KV cache costs.
Router weights are omitted, which add $\mathcal{O}(Hd_m)$ always-active parameters for head routing.
The mechanisms are complementary: GQA shares KV representations, \method{} shortens routed histories, and SA selects blocks within those histories.
With KV-group routing and group-shared block selection, combining all three stores $TK/g$ KV pairs and activates approximately $\alpha TK^2/(Hg)$ historical pairs per decoding token.
Appendix~\ref{app:complexity} details the combination rules, selection overhead, and prefill computation.

\begin{table}[t]
    \centering
    \vspace{-0.4cm}
    \caption{
        \textbf{Attention weight and KV cache costs.}
        Total and active projection weights, KV storage after $T$ tokens,
        and historical KV activation for single-token decoding.
    }
    \vspace{-0.2cm}
    \label{tab:attention-complexity}
    \begingroup
    \footnotesize
    \setlength{\tabcolsep}{3pt}
    \renewcommand{\arraystretch}{1.3}
    \resizebox{\linewidth}{!}{%
    \begin{tabular}{@{}lcccc>{\bfseries\boldmath}c|cccc@{}}
        \toprule
        Cost
        & MHA
        & GQA
        & SA
        & MoH
        & \method{}
        & \shortstack{GQA\\+SA}
        & \shortstack{GQA\\+\textbf{\method{}}}
        & \shortstack{SA\\+\textbf{\method{}}}
        & \shortstack{GQA+SA\\+\textbf{\method{}}}\\
        \midrule
        Total weights
        & $4Hd_md_h$
        & $2Hd_md_h\frac{g+1}{g}$
        & $4Hd_md_h$
        & $4Hd_md_h$
        & $4Hd_md_h$
        & $2Hd_md_h\frac{g+1}{g}$
        & $2Hd_md_h\frac{g+1}{g}$
        & $4Hd_md_h$
        & $2Hd_md_h\frac{g+1}{g}$\\
        Active weights
        & $4Hd_md_h$
        & $2Hd_md_h\frac{g+1}{g}$
        & $4Hd_md_h$
        & $4Hd_md_h$
        & $4Kd_md_h$
        & $2Hd_md_h\frac{g+1}{g}$
        & $2Kd_md_h\frac{g+1}{g}$
        & $4Kd_md_h$
        & $2Kd_md_h\frac{g+1}{g}$\\
        \midrule
        KV storage
        & $HT$
        & $HT/g$
        & $HT$
        & $HT$
        & $KT$
        & $HT/g$
        & $KT/g$
        & $KT$
        & $KT/g$\\
        KV activation
        & $HT$
        & $HT/g$
        & $\alpha HT$
        & $KT$
        & $TK^2/H$
        & $\alpha HT/g$
        & $TK^2/(Hg)$
        & $\alpha TK^2/H$
        & $\alpha TK^2/(Hg)$\\
        \bottomrule
    \end{tabular}%
    }
    \endgroup
    \vspace{-0.3cm}
\end{table}

\noindent\textbf{Head-relative positional encoding.}
We optionally apply RoPE using a token's rank within its routed head rather than its global position.
For an active token-head pair, its zero-based rank $\rho_i(t)$ and position-encoded vectors are
\begin{equation}
    \rho_i(t)=\sum_{s=0}^{t}m_{s,i}-1,\qquad
    \widetilde q_{t,i}=q_{t,i}R(\rho_i(t)),\qquad
    \widetilde k_{t,i}=k_{t,i}R(\rho_i(t)),
    \label{eq:namoh-head-rope}
\end{equation}
where $R(p)\in\mathbb{R}^{d_{\mathrm{head}}\times d_{\mathrm{head}}}$ is the RoPE rotation for row vectors at position $p$.
For example, head $0$ in Figure~\ref{fig:method-namoh}(a) maps global positions $(0,2,5,6)$ to local indices $(0,1,2,3)$.
Its next activated token, token $8$, receives index $4$.
The largest index in head $i$ is $n_i-1$, so balanced routing contracts the encoded span from $T$ positions to approximately $TK/H$.
This preserves token order, but not original token distances.
The shorter span is intended to improve query-key matching by limiting position-induced attention noise~\citep{Wertheimer2026FrayedRA,Du2026RoPEDN}, complementing the computational benefit of sparsity.
Using global-position RoPE instead amounts to replacing $\rho_i(t)$ with $t$.

\noindent\textbf{Efficient execution.}
Training and prefill pack each $(\text{batch},\text{head})$ subsequence as an independent sequence for variable-length FlashAttention~\citep{Dao2023FlashAttention2FA}.
The total packed length is fixed at $TK$ per sequence and $NK$ per training batch, although individual head lengths vary.
Load balancing helps limit this length skew.
We further use length-aware scheduling to interleave query tiles with larger and smaller causal workloads across subsequences.
This balances cumulative work across GPU multiprocessors to reduce tail idle time, without introducing cross-subsequence attention.
Decoding launches only active head queries and reads or updates only their corresponding caches.
\section{Experiments}
\label{sec:experiments}

\subsection{Experimental Setup}
\label{sec:exp-setup}

\noindent\textbf{Models and training.}
We train models from scratch with 0.6B to 1.2B parameters, covering multi-head attention (MHA), grouped-query attention (GQA), selection-based sparse attention (SA), and \method{}.
We fix the number of layers at 16, the model width at 2048, and the head width at 64, with head counts of 4, 8, 16, and 32.
Every model receives 24B tokens from FineWeb-Edu~\citep{Penedo2024TheFD}.
This budget corresponds to 20 training tokens per parameter of the largest model, following the Chinchilla scaling guideline~\citep{Hoffmann2022TrainingCL}, and is kept identical across models.
For long-context evaluation, we further post-train models on LongAlign~\citep{Bai2024LongAlignAR}.

We use AdamW with a learning rate of $0.002$, followed by linear decay to zero over the final $20\%$ of training.
Auxiliary-loss balancing uses a coefficient of $\lambda_{\mathrm{bal}}=0.001$, while loss-free balancing uses a routing-bias update rate of $\eta=0.001$.
Other optimizer hyperparameters follow the implementation defaults of the AdamW optimizer.
All experiments are conducted on NVIDIA A100 GPUs.

\noindent\textbf{Baselines and controls.}
MHA-$H$ and SA-$H$ use $H$ heads, GQA-32KV$K$ uses 32 query heads and $K$ KV heads, and \method{}-$H$A$K$ activates $K$ of $H$ heads per token.
Our SA baseline follows the block-selection paradigm~\citep{Lu2025MoBAMO,Yuan2025NativeSA}.
Each head retains its full KV history, represents each KV block by its mean-pooled keys for selection scoring, and selects a fixed fraction of the highest-scoring blocks.
We use 32-token blocks, reuse selected indices across groups of 32 queries, and additionally maintain a 128-token sliding window.
The fraction in parentheses specifies the per-head KV activation budget.
KV storage and activation are normalized to MHA-32.
Activation counts entries accessed by the main attention operation, with shared GQA entries counted once.
To isolate the gains from routed context sparsity, we equip all MHA, GQA, and SA baselines with the head gating and load-balancing mechanisms used in prior head-routing methods~\citep{Fu2026AttentionSF,Jin2024MoHMA,Qiu2025GatedAF}, with matched settings across comparisons.

\begin{table}[!t]
\centering
\vspace{-0.4cm}
\caption{
Evaluation results of models trained from scratch with different attention mechanisms and KV budgets.
KV storage (KV Stor.) and activation (KV Act.) are normalized to MHA-32.
For sparse attention (SA), the fraction in parentheses denotes the activated KV fraction per head.
}
\label{tab:main-results}
\vspace{-0.2cm}

\begingroup
\setlength{\tabcolsep}{3.2pt}
\renewcommand{\arraystretch}{1.1}

\resizebox{\linewidth}{!}{
\begin{tabular}{cccccccccccccc}
\hline
\textbf{\begin{tabular}[c]{@{}c@{}}Attention\\Mechanism\end{tabular}}
&
\textbf{\begin{tabular}[c]{@{}c@{}}KV\\Stor.\end{tabular}}
&
\textbf{\begin{tabular}[c]{@{}c@{}}KV\\Act.\end{tabular}}
&
\textbf{MMLU}
&
\textbf{GSM8K}
&
\textbf{HEval}
&
\textbf{ARC-E}
&
\textbf{ARC-C}
&
\textbf{HellaSwag}
&
\textbf{PIQA}
&
\textbf{OBQA}
&
\textbf{BoolQ}
&
\textbf{WinoG.}
&
\textbf{Average}
\\ \hline

\textbf{MHA-32}
& $1$
& $1$
& 32.79
& 5.31
& 6.10
& 65.87
& 34.13
& 45.88
& 69.37
& 37.20
& 46.64
& 53.99
& 39.73
\\ \hline

MHA-16
& $1/2$
& $1/2$
& 32.35
& 3.11
& 4.27
& 63.76
& 34.39
& 46.84
& 70.08
& 35.80
& 46.18
& 52.49
& 38.93
\\

GQA-32KV16
& $1/2$
& $1/2$
& 27.48
& 4.62
& 7.32
& 65.61
& 35.58
& 47.56
& 69.48
& 36.00
& 48.40
& 52.01
& 39.41
\\

SA-32 ($1/4$)
& $1$
& $1/4$
& 25.27
& 2.58
& 3.66
& 60.87
& 33.21
& 45.20
& 59.90
& 37.60
& 44.22
& 50.12
& 36.26
\\

SA-16 ($1/2$)
& $1/2$
& $1/4$
& 24.36
& 2.88
& 3.05
& 63.22
& 34.74
& 45.00
& 59.36
& 35.60
& 53.39
& 49.49
& 37.11
\\

\rowcolor[HTML]{EEEEEF}
\textbf{\method{}-32A16}
& $1/2$
& $1/4$
& 32.74
& 5.84
& 9.76
& 66.04
& 36.43
& 48.11
& 70.67
& 36.60
& 49.33
& 55.81
& 41.13
\\ \hline

MHA-8
& $1/4$
& $1/4$
& 26.01
& 2.58
& 3.66
& 66.04
& 33.55
& 46.02
& 69.70
& 35.20
& 44.28
& 50.80
& 37.78
\\

GQA-32KV8
& $1/4$
& $1/4$
& 25.35
& 2.88
& 6.10
& 63.72
& 32.70
& 45.49
& 69.59
& 34.60
& 48.80
& 52.75
& 38.20
\\

SA-32 ($1/16$)
& $1$
& $1/16$
& 25.12
& 1.90
& 1.83
& 61.25
& 31.84
& 42.36
& 58.60
& 26.00
& 55.17
& 50.59
& 35.46
\\

SA-8 ($1/4$)
& $1/4$
& $1/16$
& 25.11
& 1.44
& 3.05
& 61.96
& 35.32
& 43.60
& 58.87
& 29.60
& 51.90
& 50.36
& 36.12
\\

\rowcolor[HTML]{DCE3EA}
\textbf{\method{}-32A8}
& $1/4$
& $1/16$
& 31.31
& 4.62
& 7.93
& 64.86
& 36.26
& 47.78
& 69.26
& 35.80
& 55.38
& 51.54
& 40.47
\\ \hline

MHA-4
& $1/8$
& $1/8$
& 25.54
& 1.36
& 3.05
& 58.21
& 30.46
& 47.60
& 70.29
& 36.20
& 44.60
& 53.43
& 37.07
\\

GQA-32KV4
& $1/8$
& $1/8$
& 25.83
& 2.27
& 3.66
& 64.86
& 35.41
& 43.93
& 69.21
& 36.00
& 49.78
& 55.25
& 38.62
\\

SA-32 ($1/64$)
& $1$
& $1/64$
& 24.47
& 0.53
& 0.61
& 47.54
& 23.04
& 37.69
& 58.05
& 28.20
& 54.43
& 49.09
& 32.37
\\

SA-4 ($1/8$)
& $1/8$
& $1/64$
& 25.27
& 0.45
& 1.22
& 50.99
& 25.77
& 38.10
& 58.54
& 31.20
& 54.04
& 49.17
& 33.48
\\

\rowcolor[HTML]{DDE7D7}
\textbf{\method{}-32A4}
& $1/8$
& $1/64$
& 27.48
& 2.50
& 4.88
& 64.52
& 35.14
& 45.99
& 66.92
& 35.20
& 56.18
& 54.43
& 39.32
\\ \hline

\end{tabular}
}

\endgroup
\vspace{-0.1cm}
\end{table}
\begin{figure}[t]
\centering

\begin{minipage}[t]{0.46\linewidth}
\vspace{0pt}
\centering

\captionof{table}{
Evaluation results on LongBench.
KV stor. and act. are normalized to MHA-32;
SA parentheses denote the KV fraction activated.
}
\label{tab:longbench-main}

\vspace{-0.2cm}

\scriptsize
\setlength{\tabcolsep}{1.2pt}
\renewcommand{\arraystretch}{1}

\begin{tabular*}{\linewidth}{@{\extracolsep{\fill}}lcccccc@{}}
\toprule
\shortstack{\textbf{Method}\\\textbf{Dataset}}
&
\shortstack{\textbf{MHA}\\\textbf{32}}
&
\shortstack{\textbf{MHA}\\\textbf{8}}
&
\shortstack{\textbf{GQA}\\\textbf{32KV8}}
&
\shortstack{\textbf{SA-32}\\\textbf{(1/16)}}
&
\shortstack{\textbf{SA-8}\\\textbf{(1/4)}}
&
\shortstack{\textbf{\method}\\\textbf{32A8}}
\\
\midrule

\textit{KV Stor.}
& $1$
& $1/4$
& $1/4$
& $1$
& $1/4$
& $1/4$
\\

\textit{KV Act.}
& $1$
& $1/4$
& $1/4$
& $1/16$
& $1/16$
& $1/16$
\\

\midrule

HotpotQA
& 27.52 & 23.89 & \textbf{31.94} & 21.89 & 24.13 & 30.23 \\

Qasper
& 21.29 & 20.92 & 19.49 & 18.68 & 17.60 & \textbf{22.58} \\

TriviaQA
& \textbf{56.32} & 35.21 & 44.97 & 29.17 & 34.60 & 46.82 \\

NarrativeQA
& 11.14 & 12.04 & 9.93 & 9.94 & 11.06 & \textbf{13.35} \\

2WikiMQA
& 28.83 & 30.48 & 31.87 & 26.69 & 27.09 & \textbf{34.34} \\

GovReport
& \textbf{18.07} & 15.33 & 15.88 & 16.02 & 16.02 & 16.73 \\

QMSum
& 22.41 & 20.60 & 22.61 & 20.80 & 21.65 & \textbf{22.81} \\

TREC
& 71.50 & 69.50 & 70.50 & 67.00 & 68.50 & \textbf{75.00} \\

\midrule

\textbf{Average}
& 32.14 & 28.50 & 30.90 & 26.27 & 27.58 & \textbf{32.73} \\

\bottomrule
\end{tabular*}

\end{minipage}
\hfill
\begin{minipage}[t]{0.51\linewidth}
\vspace{0pt}
\centering

\includegraphics[width=\linewidth]{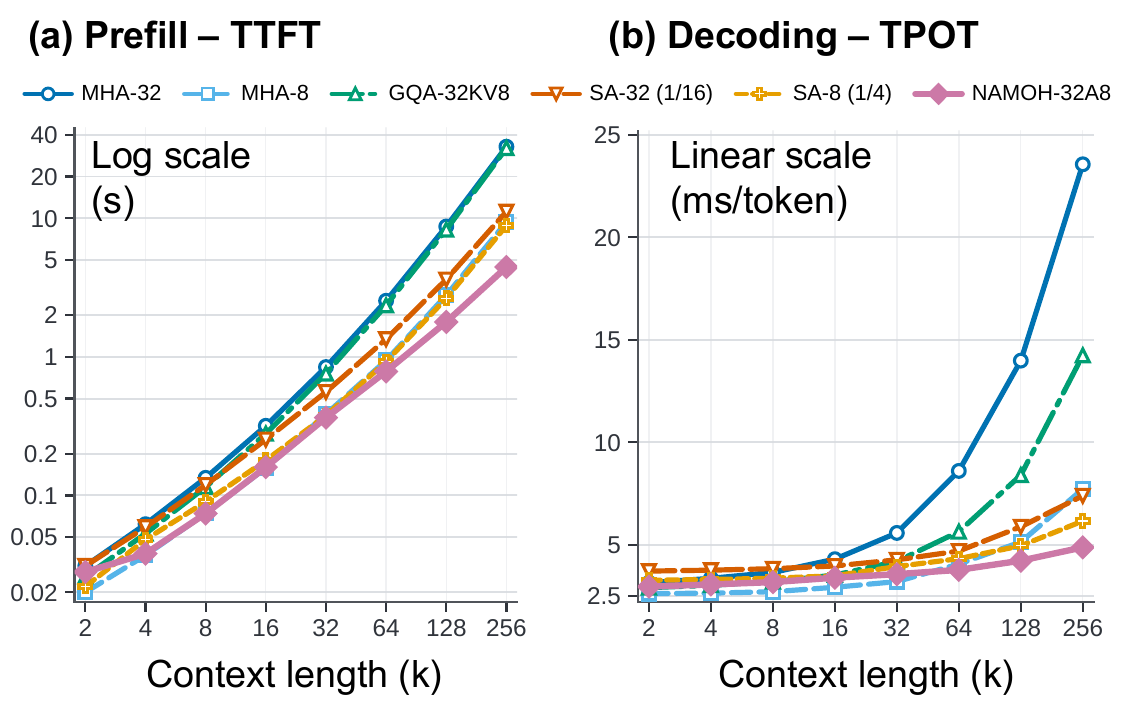}

\vspace{-0.3cm}

\captionof{figure}{
Inference efficiency.
(a) Prefill TTFT and (b) decoding TPOT across attention mechanisms at different context lengths. Lower is better.
}
\label{fig:long-efficiency}

\end{minipage}

\vspace{-0.15cm}
\end{figure}

\noindent\textbf{Evaluation.}
We evaluate pretrained models on MMLU~\citep{Hendrycks2020MeasuringMM} for general knowledge, GSM8K~\citep{Cobbe2021TrainingVT} for mathematics, HumanEval~\citep{Chen2021EvaluatingLL} for coding, and BoolQ~\citep{Clark2019BoolQET} for reading comprehension.
Scientific reasoning benchmarks include ARC-Easy, ARC-Challenge~\citep{Clark2018ThinkYH}, and OpenBookQA~\citep{Mihaylov2018CanAS}.
We assess commonsense reasoning with HellaSwag~\citep{Zellers2019HellaSwagCA}, PIQA~\citep{Bisk2019PIQARA}, and WinoGrande~\citep{Sakaguchi2019WinoGrande}.
For long-context understanding, we evaluate models on eight LongBench tasks~\citep{Bai2023LongBenchAB}: HotpotQA~\citep{Yang2018HotpotQAAD}, Qasper~\citep{Dasigi2021ADO}, TriviaQA~\citep{Joshi2017TriviaQAAL}, NarrativeQA~\citep{Kocisk2017TheNR}, 2WikiMultiHopQA~\citep{Ho2020ConstructingAM}, GovReport~\citep{Huang2021EfficientAF}, QMSum~\citep{Zhong2021QMSumAN}, and TREC~\citep{Li2002LearningQC}.

\subsection{Model Quality and Inference Efficiency}
\label{sec:exp-main}

\noindent\textbf{General capabilities.}
Table~\ref{tab:main-results} compares \method{}-32A$K$ with MHA, GQA, and SA for $K\in\{4,8,16\}$.
All variants in this table use CV-based importance regularization for load balancing, with details provided in Appendix~\ref{app:load-balancing}.
Across these settings, \method{} achieves stronger overall performance than the corresponding $K$-head MHA and SA models and compares favorably with GQA.
Its performance can also match or exceed MHA-32 despite sparse head activation.
These results show that a larger pool of selectively activated heads can preserve the quality of a fully activated model while using fewer head parameters per token.

\noindent\textbf{Long-context quality and KV budgets.}
Table~\ref{tab:longbench-main} extends the comparison to LongBench with a maximum context length of 32K tokens, where \method{}-32A8 improves overall performance over the MHA, GQA, and SA baselines.
Under balanced routing, its normalized KV storage is $1/4$ and its KV activation is approximately $1/16$.
MHA-8 and GQA-32KV8 match its storage but access approximately four times as many KV entries.
SA-32 ($1/16$) matches its activation budget but requires four times the storage, while SA-8 ($1/4$) matches both.

\begin{figure}[!t]
\centering
\vspace{-0.4cm}

\begin{minipage}[t]{0.505\textwidth}
\vspace{0pt}
\centering

\includegraphics[width=\linewidth]{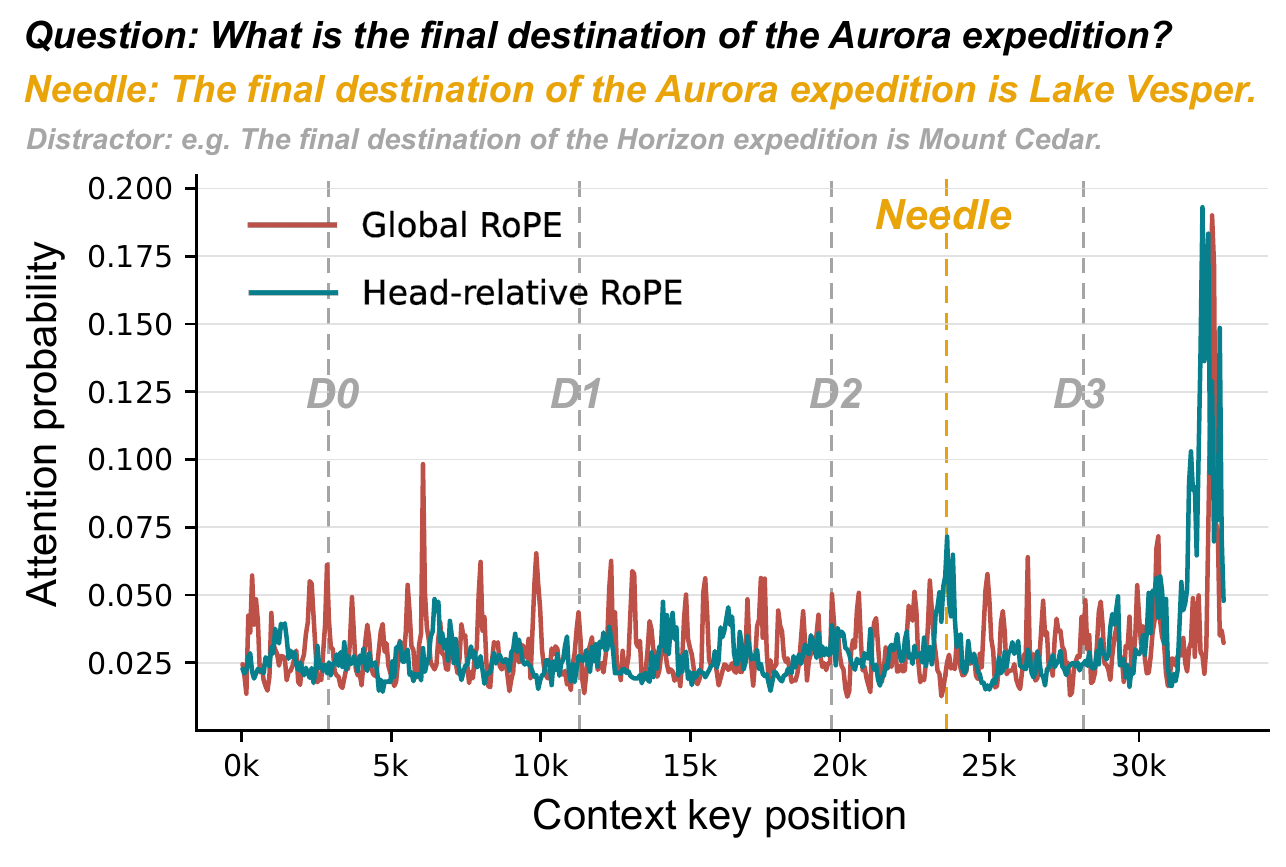}

\vspace{-0.4cm}

\captionof{figure}{
Full-question attention over a 32k-token haystack in \method{}-32A8. Dashed lines mark the needle and four distractors. 
}
\label{fig:needle-in-a-haystack}

\end{minipage}
\hfill
\begin{minipage}[t]{0.465\textwidth}
\vspace{0pt}
\centering

\captionof{table}{
LongBench evaluation of \method{}-32A8 with different RoPE variants across different context lengths.
G and HR denote global RoPE and head-relative RoPE.
}
\label{tab:longbench-rope}

\vspace{-0.15cm}

\scriptsize
\setlength{\tabcolsep}{3pt}
\renewcommand{\arraystretch}{1.05}

\begin{tabular*}{\linewidth}{@{\extracolsep{\fill}}lcccccc@{}}
\toprule

\textbf{Length}
&
\multicolumn{2}{c}{\textbf{8k}}
&
\multicolumn{2}{c}{\textbf{16k}}
&
\multicolumn{2}{c}{\textbf{32k}}
\\

\cmidrule(lr){2-3}
\cmidrule(lr){4-5}
\cmidrule(lr){6-7}

\textbf{Method}
&
\textbf{G}
&
\textbf{HR}
&
\textbf{G}
&
\textbf{HR}
&
\textbf{G}
&
\textbf{HR}
\\

\midrule

HotpotQA
&24.98&\textbf{27.46}&25.65&\textbf{29.43}&26.69&\textbf{30.23}\\

Qasper
&\textbf{19.88}&18.21&20.41&\textbf{24.12}&20.03&\textbf{22.58}\\

TriviaQA
&\textbf{45.75}&45.11&46.55&\textbf{50.10}&\textbf{47.63}&46.82\\

NarrativeQA
&9.11&\textbf{12.16}&13.26&\textbf{15.28}&9.65&\textbf{13.35}\\

2WikiMQA
&28.53&\textbf{30.54}&31.90&\textbf{32.48}&31.64&\textbf{34.34}\\

GovReport
&16.60&\textbf{18.13}&\textbf{15.65}&15.50&\textbf{17.12}&16.73\\

QMSum
&21.18&\textbf{21.49}&\textbf{23.52}&22.27&22.60&\textbf{22.81}\\

TREC
&67.00&\textbf{68.00}&\textbf{70.50}&\textbf{70.50}&73.00&\textbf{75.00}\\

\midrule

\textbf{Average}
&29.13
&\textbf{30.14}
&30.93
&\textbf{32.46}
&31.05
&\textbf{32.73}
\\

\bottomrule

\end{tabular*}

\end{minipage}

\end{figure}
\begin{figure}[!t]
    \centering
    \vspace{-0.4cm}
    \includegraphics[width=\linewidth]{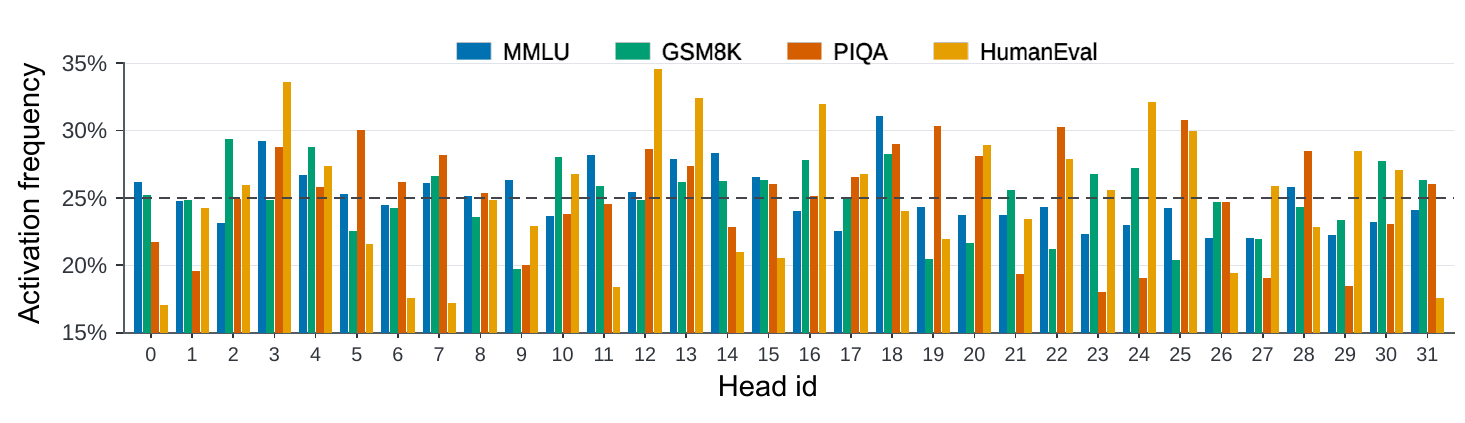}
    \vspace{-0.8cm}
    \caption{
        Task-dependent head utilization in \method{}-32A8.
        Activation frequencies of individual heads in one attention layer on MMLU, GSM8K, PIQA, and HumanEval.
    }
    \label{fig:head-load}
    \vspace{-0.4cm}
\end{figure}

\noindent\textbf{Inference efficiency.}
Figure~\ref{fig:long-efficiency} reports time to first token (TTFT) and time per output token (TPOT) across context lengths.
TTFT measures the latency to process the prompt and produce the first output token.
TPOT measures the average latency of subsequent decoding steps.
For the SA baselines, we use the implementation from MoBA~\citep{Lu2025MoBAMO}.
At long contexts, \method{}-32A8 achieves lower TTFT and TPOT than MHA-32, GQA-32KV8, and SA variants, and even outperforms MHA-8.

These gains are consistent with the different bottlenecks of decoding and prefill.
Long-context decoding is typically memory-bound, so reducing KV traffic helps lower TPOT~\citep{Yuan2025NativeSA}.
Compute-bound prefill benefits from fewer query-key interactions and sparse projection.
GQA reduces KV projection costs but retains full-prefix attention for all 32 query heads.
SA reduces the attended KV set but still incurs block-scoring and selection overhead~\citep{Lu2025MoBAMO,Yuan2025NativeSA}.
In contrast, \method{} performs attention over shorter packed subsequences without a history-scanning indexer, which also benefits comparisons at matched KV activation.

\subsection{Positional Encoding and Routing Behavior}
\label{sec:exp-analysis}

\noindent\textbf{Head-relative RoPE.}
Figure~\ref{fig:needle-in-a-haystack} compares global and head-relative RoPE in \method{}-32A8 on a 32k-token needle-in-a-haystack example with four distractors.
The visualization aggregates attention from all question tokens to each haystack position.
In this example, head-relative RoPE assigns more attention to the needle and less to irrelevant positions.
Table~\ref{tab:longbench-rope} provides a broader comparison at context lengths of 8k, 16k, and 32k.
Head-relative RoPE achieves higher LongBench scores, with the average gap increasing at longer contexts.
These observations are consistent with the intended benefit of shortening the encoded positional span, as discussed in Section~\ref{sec:namoh-properties}.

\noindent\textbf{Head utilization across tasks.}
Figure~\ref{fig:head-load} shows head activation frequencies on four datasets.
Frequencies cluster around the balanced rate of $K/H=25\%$, indicating broad head utilization rather than concentration on a small fixed subset.
At the same time, individual heads exhibit clear frequency differences across tasks.
This pattern suggests task-dependent specialization.

\begin{figure}[t]
    \centering
    \vspace{-0.6cm}
    \includegraphics[width=\linewidth]{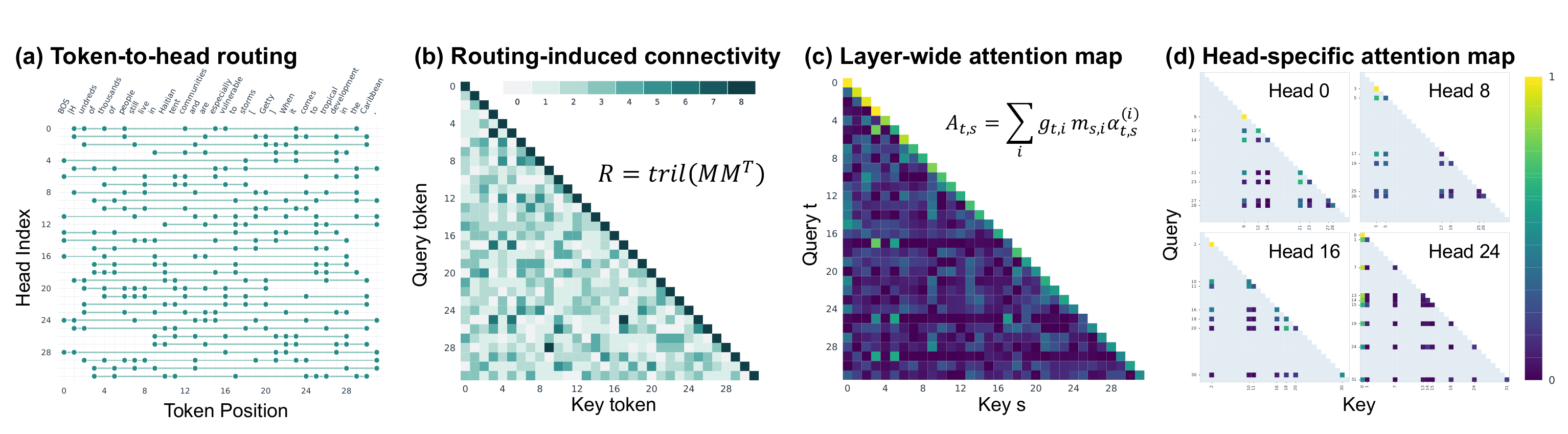}
    \vspace{-0.6cm}
    \caption{
        Routing shapes attention connectivity in \method{}.
        (a) Token-to-head assignments.
        (b) The number of shared active heads for each causal token pair.
        (c) Layer-wide attention weighted by query-specific routing gates.
        (d) Attention maps for heads 0, 8, 16, and 24.
    }
    \vspace{-0.1cm}
    \label{fig:attention-map}
\end{figure}

\noindent\textbf{Routing-induced attention structure.}
Figure~\ref{fig:attention-map} examines a 32-token example.
Panel (a) shows the heads selected by each token, with lines linking tokens assigned to the same head.
For this sequence of length $T=32$, let $M\in\{0,1\}^{T\times H}$ collect the routing masks, with $M_{t,i}=m_{t,i}$.
Panel (b) visualizes $R=\operatorname{tril}(MM^\top)$, where $\operatorname{tril}$ retains the lower triangle, including the diagonal.
Thus, $R_{t,s}$ counts the active heads shared by query token $t$ and an earlier or current token $s$.

Panel (c) shows the gate-weighted attention map $A_{t,s}=\sum_{i=0}^{H-1}g_{t,i}m_{s,i}\alpha_{t,s}^{(i)}$.
Here, $\alpha_{t,s}^{(i)}$ is the softmax attention weight from token $t$ to token $s$ within head $i$, defined as zero outside that head's routed causal pairs.
Panels (b) and (c) therefore distinguish available connections from the attention weights assigned to them.
Panel (d) displays individual head maps at the original token positions.
Their separated support reflects the different routed subsequences, with non-routed positions absent from each head's computation.

\subsection{Load Balancing and Shared Heads}
\label{sec:exp-ablations}

\noindent\textbf{Load-balancing strategies.}
Table~\ref{tab:namoh-balancing-shared} compares CV-based importance regularization~\citep{Shazeer2017OutrageouslyLN}, Switch-style ($fp$) balancing~\citep{Fedus2021SwitchTS}, and loss-free balancing~\citep{Wang2024AuxiliaryLossFreeLB}.
The formulations of these load-balancing strategies are detailed in Appendix~\ref{app:load-balancing}.
All variants retain 32 routed heads and activate eight per token.
At the settings specified above, $fp$ achieves the best overall performance among the three alternatives.

\begin{table}[!t]
\centering
\caption{ 
Further refinement of \method{}-32A8 with load-balancing strategies and shared heads.
}
\label{tab:namoh-balancing-shared}
\vspace{-0.2cm}
\begingroup
\setlength{\tabcolsep}{3.2pt}
\renewcommand{\arraystretch}{1.1}
\resizebox{\linewidth}{!}{
\begin{tabular}{lccccccccccc}
\hline
\textbf{Method}
& \textbf{MMLU}
& \textbf{GSM8K}
& \textbf{HEval}
& \textbf{ARC-E} 
& \textbf{ARC-C}
& \textbf{HellaSwag}
& \textbf{PIQA}
& \textbf{OBQA}
& \textbf{BoolQ}
& \textbf{WinoG.}
& \textbf{Average} \\ \hline

\multicolumn{12}{l}{
{\color[HTML]{656565}
\textit{\textbf{Load Balancing Variants}}}
} \\

\textbf{CV}~\citep{Shazeer2017OutrageouslyLN}
& 31.31 & 4.62 & 7.93 & 64.86 & 36.26
& 47.78 & 69.26 & 35.80 & 55.38 & 51.54
& 40.47 \\

\cellcolor[HTML]{DCE3EA}\textbf{fp}~\citep{Fedus2021SwitchTS}
& \cellcolor[HTML]{DCE3EA}32.40
& \cellcolor[HTML]{DCE3EA}5.31
& \cellcolor[HTML]{DCE3EA}9.76
& \cellcolor[HTML]{DCE3EA}64.66
& \cellcolor[HTML]{DCE3EA}37.11
& \cellcolor[HTML]{DCE3EA}46.16
& \cellcolor[HTML]{DCE3EA}70.40
& \cellcolor[HTML]{DCE3EA}36.80
& \cellcolor[HTML]{DCE3EA}56.81
& \cellcolor[HTML]{DCE3EA}52.96
& \cellcolor[HTML]{DCE3EA}41.24 \\

\textbf{Loss-Free}~\citep{Wang2024AuxiliaryLossFreeLB}
& 33.23 & 3.11 & 7.93 & 62.76 & 34.23
& 45.12 & 68.23 & 35.00 & 55.63 & 52.96
& 39.82 \\ \hline

\multicolumn{12}{l}{
{\color[HTML]{656565}
\textit{\textbf{Shared-Head Variants}}}
} \\

\textbf{+ 2 Shared (Full)}
& 30.99 & 5.38 & 8.64 & 64.10 & 37.64
& 47.12 & 70.97 & 37.00 & 58.31 & 53.28
& 41.34 \\

\cellcolor[HTML]{DDE7D7}\textbf{+ 2 Shared (Sliding)}
& \cellcolor[HTML]{DDE7D7}32.21
& \cellcolor[HTML]{DDE7D7}5.53
& \cellcolor[HTML]{DDE7D7}9.76
& \cellcolor[HTML]{DDE7D7}65.19
& \cellcolor[HTML]{DDE7D7}37.21
& \cellcolor[HTML]{DDE7D7}47.51
& \cellcolor[HTML]{DDE7D7}70.88
& \cellcolor[HTML]{DDE7D7}36.80
& \cellcolor[HTML]{DDE7D7}57.02
& \cellcolor[HTML]{DDE7D7}52.33
& \cellcolor[HTML]{DDE7D7}41.44 \\

\textbf{+ 4 Shared (Full)}
& 33.14 & 6.22 & 10.37 & 66.12 & 38.24
& 47.37 & 71.24 & 37.60 & 58.63 & 53.75
& 42.27 \\

\cellcolor[HTML]{DDE7D7}\textbf{+ 4 Shared (Sliding)}
& \cellcolor[HTML]{DDE7D7}33.50
& \cellcolor[HTML]{DDE7D7}5.84
& \cellcolor[HTML]{DDE7D7}11.59
& \cellcolor[HTML]{DDE7D7}66.33
& \cellcolor[HTML]{DDE7D7}38.75
& \cellcolor[HTML]{DDE7D7}48.74
& \cellcolor[HTML]{DDE7D7}72.06
& \cellcolor[HTML]{DDE7D7}38.40
& \cellcolor[HTML]{DDE7D7}59.86
& \cellcolor[HTML]{DDE7D7}54.22
& \cellcolor[HTML]{DDE7D7}42.93 \\ \hline

\end{tabular}
}
\vspace{-0.3cm}
\endgroup
\end{table}

\noindent\textbf{Shared heads for local context.}
Inspired by shared expert isolation in DeepSeekMoE~\citep{Dai2024DeepSeekMoETU}, we augment \method{}-32A8 with $H_s$ always-active shared heads, testing $H_s\in\{2,4\}$.
Each shared head attends either to the full causal prefix or to the most recent $w=128$ original tokens, including the current token, where $w$ is the sliding-window size.
Their independently projected outputs are added to the routed output $o_t$.
Shared heads are excluded from $H$, $K$, and the routing balance statistics.
The sliding-window heads use global RoPE, while routed heads retain head-relative RoPE.
This supplies local context even when neighboring tokens select different routed heads.

Shared heads add $4H_sd_{\mathrm{model}}d_{\mathrm{head}}$ always-active projection parameters.
For a fixed window size $w$, the sliding variant requires at most $H_sw$ additional KV pairs per layer and $\mathcal{O}(TH_sw d_{\mathrm{head}})$ attention work for a length-$T$ sequence.
Table~\ref{tab:namoh-balancing-shared} shows that adding shared heads improves overall performance.
Sliding-window variants perform similarly to their full-attention counterparts, suggesting that reliable local context accounts for much of the benefit.
This supports a complementary design in which shared heads cover nearby tokens while routed heads can focus on broader context retrieval.
\section{Conclusion}
\label{sec:conclusion}

We introduced \method{}, an architecture-native sparse attention mechanism that connects attention parameter scaling with context scaling.
Each token activates a subset of heads, and each head stores and attends only to its assigned tokens.
Routing thus jointly selects active parameters and available context without scanning the full history.
Under balanced routing, expanding the head pool at a fixed active head count shortens head histories.
This reduces per-token KV access without increasing total KV storage.
Head-relative RoPE further shortens positional spans and improves long-context quality in our evaluations.
Experiments show that \method{} can outperform fully activated models with the same total parameter count.
It also enables faster long-context prefill and decoding than smaller dense models matched in active parameter count.
The design remains compatible with GQA and existing sparse attention mechanisms.
Together, these findings support a complementary path for scaling attention, in which parameter growth directly enables more efficient and effective context scaling.

\newpage


\bibliography{main}
\bibliographystyle{iclr2027_conference}


\newpage
\appendix

\section{Load Balancing Strategies}
\label{app:load-balancing}

We describe three balancing strategies for a single routed attention layer.
For a training batch $\mathcal{B}$ containing $N$ tokens, each token selects $K$ of $H$ heads.
We use the head affinities $a_{t,i}$, selection indicators $m_{t,i}$, assignment fractions $f_i$, and average affinities $p_i$ defined in the main text.
Auxiliary losses are summed across routed attention layers and weighted by $\lambda_{\mathrm{bal}}$.
Shared heads are excluded from all balancing statistics.

\subsection{CV-Based Load Balancing}
\label{app:load-balancing-cv}

Following \citet{Shazeer2017OutrageouslyLN}, CV-based balancing penalizes variation in both routing importance and expected token load.
For a nonnegative head-statistic vector $\boldsymbol{u}=(u_0,\ldots,u_{H-1})$ with positive mean $\bar u$, the squared coefficient of variation is
\begin{equation}
    \operatorname{CV}(\boldsymbol{u})^2
    =\frac{H^{-1}\sum_{i=0}^{H-1}(u_i-\bar u)^2}{\bar u^2},
    \qquad
    \bar u=\frac{1}{H}\sum_{i=0}^{H-1}u_i.
    \label{eq:namoh-cv-definition}
\end{equation}
This measures variance relative to the squared mean and is minimized when all entries are equal.

For head $i$, define its routing importance $I_i$ and expected assignment count $\ell_i$ as
\begin{equation}
    I_i=\sum_{t\in\mathcal{B}}a_{t,i}=Np_i,
    \qquad
    \ell_i=\sum_{t\in\mathcal{B}}
    \Pr_{\mathrm{noise}}\!\left(i\in\mathcal{S}_K^{\mathrm{noise}}(t)\right),
    \label{eq:namoh-cv-statistics}
\end{equation}
where $\mathcal{S}_K^{\mathrm{noise}}(t)$ denotes the heads selected by noisy Top-$K$ routing, and the probability is taken over the routing noise.
The importance term measures total affinity, while the expected-load term measures how often a head is selected under noisy routing.
Let $\boldsymbol{I}=(I_0,\ldots,I_{H-1})$ and $\boldsymbol{\ell}=(\ell_0,\ldots,\ell_{H-1})$.
The combined loss is
\begin{equation}
    \mathcal{L}_{\mathrm{CV}}
    =\alpha_{\mathrm{imp}}\operatorname{CV}(\boldsymbol{I})^2
    +\alpha_{\mathrm{load}}\operatorname{CV}(\boldsymbol{\ell})^2,
    \label{eq:namoh-cv-balancing}
\end{equation}
where $\alpha_{\mathrm{imp}},\alpha_{\mathrm{load}}>0$ set the relative weights of the two penalties.
Their overall strength is controlled by $\lambda_{\mathrm{bal}}$.

The importance penalty is equivalent to $H\sum_i(p_i-1/H)^2$ and is differentiable through the head affinities.
The load penalty additionally requires a differentiable expected-load estimator, obtained through noisy Top-$K$ routing~\citep{Shazeer2017OutrageouslyLN}.
It is not computed by directly differentiating the hard assignment counts $NKf_i$.
Thus, the full CV formulation balances both affinity mass and expected head utilization, rather than affinity mass alone.

\subsection{Switch-Style Auxiliary Balancing}
\label{app:load-balancing-switch}

The Switch-style loss~\citep{Fedus2021SwitchTS} combines observed head utilization with differentiable routing affinities:
\begin{equation}
    \mathcal{L}_{fp}
    =H\sum_{i=0}^{H-1}f_ip_i.
    \label{eq:namoh-fp-balancing}
\end{equation}
For Top-$K$ head routing, $f_i$ is normalized by the total number of assignments $NK$, so both $\{f_i\}$ and $\{p_i\}$ sum to one.
The loss equals one under uniform assignments and affinities, independently of $H$ and $K$.

During backpropagation, $f_i$ is treated as constant, while gradients flow through $p_i$.
Since $\partial\mathcal{L}_{fp}/\partial p_i=Hf_i$, a head with a larger assignment fraction receives a stronger penalty on its average affinity.
The observed load therefore guides the router toward less-used heads without requiring a differentiable estimate of assignment counts.

This objective is also related to CV regularization.
For $\boldsymbol{f}=(f_0,\ldots,f_{H-1})$, when $p_i\approx f_i$,
\begin{equation}
    \mathcal{L}_{fp}
    \approx H\sum_{i=0}^{H-1}f_i^2
    =1+\operatorname{CV}(\boldsymbol{f})^2.
    \label{eq:namoh-fp-cv-relation}
\end{equation}
Unlike direct count-based CV regularization, the affinity factor provides a differentiable path to the router.
As with the CV strategy, an overly large $\lambda_{\mathrm{bal}}$ can make balancing gradients compete with the head specialization favored by the language-modeling objective.

\subsection{Auxiliary-Loss-Free Balancing}
\label{app:load-balancing-loss-free}

Following \citet{Wang2024AuxiliaryLossFreeLB}, loss-free balancing controls head utilization through routing biases rather than an auxiliary objective.
Each head maintains a scalar bias $b_i$, initialized to zero.
Selection uses biased affinities, while output gates retain the original affinities:
\begin{equation}
    \mathcal{S}_K(t)
    =\operatorname{TopK}_{i}(a_{t,i}+b_i),
    \qquad
    m_{t,i}=\mathbf{1}\{i\in\mathcal{S}_K(t)\},
    \qquad
    g_{t,i}=m_{t,i}a_{t,i}.
    \label{eq:namoh-biased-routing}
\end{equation}
Thus, biases change which heads are selected but do not enter their gating weights.
Selected affinities are not renormalized.

Let $c_i$ denote the number of token assignments received by head $i$ in the current batch, and let $\bar c$ be the average count across heads.
After each completed training batch, we update
\begin{equation}
    c_i=\sum_{t\in\mathcal{B}}m_{t,i}=NKf_i,
    \qquad
    \bar c=\frac{NK}{H},
    \qquad
    b_i\leftarrow b_i+\eta\,\operatorname{sign}(\bar c-c_i),
    \label{eq:namoh-loss-free}
\end{equation}
where $\eta>0$ is the update rate and $\operatorname{sign}(0)=0$.
An overloaded head receives a negative bias adjustment, reducing its chance of future selection.
An underloaded head receives a positive adjustment.
This forms a feedback loop from observed assignments to subsequent routing decisions.

Bias updates occur outside backpropagation.
In \method{}, biases remain fixed within each training batch and during inference, so updates do not revise earlier token assignments.
The router learns token-to-head affinities through the language-modeling objective, while the biases regulate head utilization.
Unlike the two auxiliary-loss strategies, this method introduces no direct load-balancing gradient into the router.
\section{Complexity Analysis and Comparisons}
\label{app:complexity}

\subsection{Setup and Accounting}
\label{app:complexity-setup}

We analyze one layer with $H$ query heads, model width $d_m=d_{\mathrm{model}}$, head width $d_h=d_{\mathrm{head}}$, and a prefix of $T\geq1$ tokens.
Routed variants activate $K$ query heads per token, with $1\leq K\leq H$.
GQA places $g$ query heads in each KV group, where $g\mid H$; group-routed variants additionally require $g\mid K$.
For ungrouped variants, $g=1$.
SA uses an integer block size $B\geq1$ and a retained fraction $\alpha\in(0,1]$ of each available local history.
All widths are held fixed; we do not require $d_m=Hd_h$.

Let $P_{\mathrm{total}}$ and $P_{\mathrm{active}}$ count all learned weights and those evaluated for one token.
We use a single bias-free linear router for MoH and \method{} to isolate their attention differences.
An $H$-way head router contains $Hd_m$ weights, while an $H/g$-way group router contains $Hd_m/g$.
All router weights are active for every token.
Projection biases, always-active shared heads, and extra attention branches are excluded.
The metadata-based block selectors considered here introduce no learned weights.

As in the main text, $M_{\mathrm{KV}}$ counts stored KV pairs, and $A_{\mathrm{KV}}(T)$ counts distinct historical pairs used by the next token.
Multiplying either count by $2d_h$ gives its scalar data volume.
Let $C_{\mathrm{attn}}$ count all causal query-key interactions during prefill, including self-attention.
Unlike KV activation, this count includes separate interactions for query heads that share the same KV pair.
All counts are logical and exclude padding, temporary buffers, and training activations.

\subsection{MHA and GQA}
\label{app:complexity-dense}

\noindent\textbf{MHA.}
Each head has independent query, key, value, and output projections~\citep{Vaswani2017AttentionIA}.
Each projection contributes $d_md_h$ weights, so
\begin{equation}
    P_{\mathrm{total}}=P_{\mathrm{active}}=4Hd_md_h,\qquad
    M_{\mathrm{KV}}=HT,\qquad
    A_{\mathrm{KV}}(T)=HT.
\end{equation}
A query at one-based causal rank $r$ attends to $r$ entries per head.
Summing over all ranks gives
\begin{equation}
    C_{\mathrm{attn}}
    =H\sum_{r=1}^{T}r
    =\frac{HT(T+1)}{2}.
    \label{eq:app-mha-prefill}
\end{equation}

\noindent\textbf{GQA.}
Let $\mathcal{Q}_j$ denote the $g$ query heads assigned to KV group $j$, for $j=0,\ldots,H/g-1$.
GQA retains separate query and output projections but shares key and value projections within each group~\citep{Ainslie2023GQATG}.
Therefore,
\begin{equation}
\begin{aligned}
    P_{\mathrm{total}}=P_{\mathrm{active}}
    &=2Hd_md_h+\frac{2H}{g}d_md_h
      =2H\left(1+\frac1g\right)d_md_h,\\
    M_{\mathrm{KV}}&=\frac{HT}{g},\qquad
    A_{\mathrm{KV}}(T)=\frac{HT}{g}.
\end{aligned}
\end{equation}
Each stored key still interacts with $g$ queries at each causal position:
\begin{equation}
    C_{\mathrm{attn}}
    =\sum_{j=0}^{H/g-1}\sum_{r=1}^{T}gr
    =\frac{HT(T+1)}{2}.
    \label{eq:app-gqa-prefill}
\end{equation}
Thus, KV sharing reduces storage and distinct access, but not the query-key interaction count at fixed $H$.

\subsection{Block-Selected Sparse Attention}
\label{app:complexity-selection}

\noindent\textbf{Selection model.}
We analyze a fixed-fraction version of block selection that scans historical metadata, using blocks of $B$ entries.
MoBA scores mean-pooled keys, while Quest uses channelwise key minima and maxima~\citep{Lu2025MoBAMO,Tang2024QuestQS}.
We apply the specified selector during both prefill and decoding; this does not reproduce every cited method's execution settings.

For a nonnegative history length $\ell$, let $N_B(\ell)$ be its number of blocks.
At local causal rank $r\geq1$, let $R_{\alpha,B}(r)$ be the selected block count, including the current block:
\begin{equation}
    N_B(\ell)=\left\lceil\frac{\ell}{B}\right\rceil,
    \qquad
    R_{\alpha,B}(r)=\left\lceil\alpha N_B(r)\right\rceil.
    \label{eq:app-block-budget}
\end{equation}
For example, $\alpha=0.2$ targets approximately $20\%$ of each local history.
We always include the current block and select the remaining blocks from completed historical blocks.
Future blocks are excluded.
The current block uses a causal mask and is not ranked using metadata that contains future keys, following the causality rule of MoBA~\citep{Lu2025MoBAMO}.

\noindent\textbf{Exact selected-entry count.}
Let $a_{\alpha,B}(r)$ count entries attended by a query at local rank $r$, including itself.
Each selected historical block contains $B$ entries, and the current block contributes $r-B[N_B(r)-1]$ visible entries.
Hence,
\begin{equation}
\begin{aligned}
    a_{\alpha,B}(r)
    &=B[R_{\alpha,B}(r)-1]+r-B[N_B(r)-1]\\
    &=r-B[N_B(r)-R_{\alpha,B}(r)].
\end{aligned}
\label{eq:app-selected-entries}
\end{equation}
Writing $\varepsilon_r=a_{\alpha,B}(r)-\alpha r$ for the rounding error gives
\begin{equation}
\begin{aligned}
    \varepsilon_r
    &=B[R_{\alpha,B}(r)-\alpha N_B(r)]
      +(1-\alpha)[r-BN_B(r)],\\
    |\varepsilon_r|&\leq B.
\end{aligned}
\label{eq:app-block-rounding}
\end{equation}
Thus, $a_{\alpha,B}(r)=\alpha r+\mathcal{O}(B)$.
The relative approximation is useful when $\alpha r\gg B$; a history contained in one block is attended densely.

SA retains every KV pair because a later query may select a previously unused block.
Its weights are unchanged from MHA, and
\begin{equation}
\begin{aligned}
    M_{\mathrm{KV}}&=HT,\\
    A_{\mathrm{KV}}(T)
    &=H[a_{\alpha,B}(T+1)-1]
      =\alpha HT+\mathcal{O}(HB).
\end{aligned}
\label{eq:app-sa-kv}
\end{equation}
The subtraction removes the next token's self-entry.

\noindent\textbf{Causal prefill interactions.}
Define the selected causal sum for a local sequence of length $\ell$ as
\begin{equation}
\begin{aligned}
    \Phi_{\alpha,B}(\ell)
    &=\sum_{r=1}^{\ell}a_{\alpha,B}(r)\\
    &=\frac{\ell(\ell+1)}{2}
      -B\sum_{r=1}^{\ell}[N_B(r)-R_{\alpha,B}(r)]\\
    &=\frac{\alpha\ell(\ell+1)}{2}
      +\sum_{r=1}^{\ell}\varepsilon_r\\
    &=\frac{\alpha\ell(\ell+1)}{2}+\mathcal{O}(B\ell).
\end{aligned}
\label{eq:app-fractional-causal-sum}
\end{equation}
Self-attention and the mandatory current block are included exactly in $\Phi_{\alpha,B}$.
Applying this sum to all $H$ heads yields
\begin{equation}
    C_{\mathrm{attn}}
    =H\Phi_{\alpha,B}(T)
    =\frac{\alpha HT^2}{2}+\mathcal{O}(HBT).
    \label{eq:app-sa-prefill}
\end{equation}

\noindent\textbf{Metadata storage and scanning.}
Let $\nu$ be the number of metadata vectors per block, each of width $d_h$.
Mean pooling uses $\nu=1$, and minima and maxima use $\nu=2$.
The metadata storage in scalars is $M_{\mathrm{meta}}=\nu d_hHN_B(T)$.
Construction costs $\mathcal{O}(HTd_h)$ and supports incremental updates.

Let $D_{\mathrm{meta}}(T)$ and $C_{\mathrm{meta}}$ count candidate block summaries inspected during the next decoding step and throughout prefill, respectively.
A query at rank $r$ scans $N_B(r)-1$ completed blocks.
To sum these visits, define
\begin{equation}
    \Gamma_B(\ell)=\sum_{r=1}^{\ell}N_B(r).
\end{equation}
Write $\ell=uB+v$, where $u=\lfloor\ell/B\rfloor$ and $0\leq v<B$.
Each complete block contributes $B$ queries with the same block count, giving
\begin{equation}
\begin{aligned}
    \Gamma_B(\ell)
    &=B\sum_{b=1}^{u}b+(u+1)v\\
    &=\frac{Bu(u+1)}{2}+(u+1)v
     =\frac{\ell^2}{2B}+\mathcal{O}(\ell).
\end{aligned}
\label{eq:app-block-scan-sum}
\end{equation}
Here, $b$ indexes complete local blocks.
Consequently,
\begin{equation}
\begin{aligned}
    D_{\mathrm{meta}}(T)
    &=H[N_B(T+1)-1]
     =H\left\lfloor\frac{T}{B}\right\rfloor,\\
    C_{\mathrm{meta}}
    &=H[\Gamma_B(T)-T]
     =\frac{HT^2}{2B}+\mathcal{O}(HT).
\end{aligned}
\label{eq:app-sa-metadata}
\end{equation}
The subtraction excludes the unscored current block.
These visits do not acquire a factor of $\alpha$, since the selector scans all candidates before choosing blocks.

\subsection{MoH}
\label{app:complexity-moh}

We use the dense-projection MoH baseline: all query, key, value, and output projection weights are evaluated, but only $K$ heads perform attention~\citep{Jin2024MoHMA}.
Every head stores every prefix token, including tokens for which its output was inactive.
Under the router convention in Section~\ref{app:complexity-setup},
\begin{equation}
\begin{aligned}
    P_{\mathrm{total}}=P_{\mathrm{active}}&=4Hd_md_h+Hd_m,\\
    M_{\mathrm{KV}}&=HT,\qquad
    A_{\mathrm{KV}}(T)=KT.
\end{aligned}
\end{equation}
At global causal rank $r$, each of the $K$ selected heads attends to all $r$ entries.
Therefore,
\begin{equation}
    C_{\mathrm{attn}}
    =\sum_{r=1}^{T}Kr
    =\frac{KT(T+1)}{2}.
\end{equation}
Head selection reduces KV activation and attention interactions, but not projection activation or cache storage in this baseline.

\subsection{\method{}}
\label{app:complexity-namoh}

\noindent\textbf{Parameters and routed storage.}
Routing precedes projection, so inactive token-head pairs produce no queries, keys, or values.
Only selected output slices are evaluated:
\begin{equation}
    P_{\mathrm{total}}=4Hd_md_h+Hd_m,\qquad
    P_{\mathrm{active}}=4Kd_md_h+Hd_m.
\end{equation}
Let $m_{t,i}$ indicate whether token $t$ selects head $i$, and let $\mathcal{S}_K(t)$ be the selected set.
The head lengths satisfy
\begin{equation}
    n_i=\sum_{t=0}^{T-1}m_{t,i},\qquad
    \sum_{i=0}^{H-1}n_i
    =\sum_{t=0}^{T-1}\sum_{i=0}^{H-1}m_{t,i}
    =TK.
    \label{eq:app-assignment-count}
\end{equation}
Only these assignments create KV pairs. Hence,
\begin{equation}
    M_{\mathrm{KV}}=TK,\qquad
    A_{\mathrm{KV}}(T)=\sum_{i\in\mathcal{S}_K(T)}n_i.
    \label{eq:app-routed-kv}
\end{equation}
Let $\bar n=TK/H$ denote the mean head length.
Under equal lengths, activation is $K\bar n=TK^2/H$.
Equal lengths require integer $\bar n$; approximate balance gives the main-text estimate.

\noindent\textbf{Causal interaction count.}
Within head $i$, the query at local rank $r$ attends to the first $r$ entries.
Thus,
\begin{equation}
\begin{aligned}
    C_{\mathrm{attn}}
    &=\sum_{i=0}^{H-1}\sum_{r=1}^{n_i}r
      =\frac12\sum_{i=0}^{H-1}n_i(n_i+1)\\
    &=\frac12\sum_{i=0}^{H-1}n_i^2+\frac{TK}{2}.
\end{aligned}
\label{eq:app-routed-interactions}
\end{equation}
For the load vector $\boldsymbol n=(n_0,\ldots,n_{H-1})$, define its squared coefficient of variation using the population variance:
\begin{equation}
    \operatorname{CV}(\boldsymbol n)^2
    =\frac{H^{-1}\sum_{i=0}^{H-1}(n_i-\bar n)^2}{\bar n^2}.
    \label{eq:app-load-cv}
\end{equation}
Since $\sum_i(n_i-\bar n)=0$, expanding around the mean gives
\begin{equation}
\begin{aligned}
    \sum_{i=0}^{H-1}n_i^2
    &=H\bar n^2+2\bar n\sum_{i=0}^{H-1}(n_i-\bar n)
      +\sum_{i=0}^{H-1}(n_i-\bar n)^2\\
    &=H\bar n^2[1+\operatorname{CV}(\boldsymbol n)^2]\\
    &=\frac{T^2K^2}{H}[1+\operatorname{CV}(\boldsymbol n)^2].
\end{aligned}
\label{eq:app-load-second-moment}
\end{equation}
Substitution into Equation~\ref{eq:app-routed-interactions} yields
\begin{equation}
    C_{\mathrm{attn}}
    =\frac{T^2K^2}{2H}[1+\operatorname{CV}(\boldsymbol n)^2]
      +\frac{TK}{2}.
    \label{eq:namoh-attention-work}
\end{equation}
The causal sum includes self-attention exactly; the second moment determines the leading cost.

\noindent\textbf{Balance and matched comparisons.}
The inequalities $\sum_i n_i^2\geq(\sum_i n_i)^2/H$ and $n_i^2\leq Tn_i$ imply
\begin{equation}
    \frac{T^2K^2}{2H}+\frac{TK}{2}
    \leq C_{\mathrm{attn}}
    \leq\frac{KT(T+1)}{2}.
    \label{eq:app-interaction-bounds}
\end{equation}
Equal lengths attain the lower bound when feasible.
If every token selects the same $K$ heads, the upper bound is attained and decoding activation becomes $KT$.
Storage remains $TK$ in both cases.
Batch-level load balancing does not guarantee balanced histories within every prefix.

For an all-active baseline with $J\in\{H,K\}$ heads, the balanced finite-length prefill ratio is
\begin{equation}
    \frac{C_{\mathrm{attn}}}{JT(T+1)/2}
    =\frac{TK^2/H+K}{J(T+1)}
    \longrightarrow\frac{K^2}{HJ}
    \quad\text{as }T\to\infty.
\end{equation}
The choices $J=H$ and $J=K$ match total and active head-projection parameters, respectively; router weights are additional.

\subsection{GQA+SA}
\label{app:complexity-gqa-sa}

\noindent\textbf{Independent or shared selection.}
Query heads sharing a KV group can select blocks independently.
Let $\mathcal{E}_{t,i}$ be the set of historical KV entries selected by query head $i$ for token $t$.
Distinct access within group $j$ then counts the union:
\begin{equation}
    \max_{i\in\mathcal Q_j}|\mathcal E_{T,i}|
    \leq\left|\bigcup_{i\in\mathcal Q_j}\mathcal E_{T,i}\right|
    \leq\min\!\left\{T,\sum_{i\in\mathcal Q_j}|\mathcal E_{T,i}|\right\}.
\end{equation}
Ignoring rounding, the union can range from $\alpha T$ to $\min(T,g\alpha T)$ entries.
We instead use one shared block selection per group, following the group-consistent design of \citet{Yuan2025NativeSA}.
Queries share block indices, not their attention weights.

\noindent\textbf{Costs with shared indices.}
Parameter counts remain those of GQA.
Each of its $H/g$ groups stores one full history and selects one block set:
\begin{equation}
\begin{aligned}
    M_{\mathrm{KV}}&=\frac{HT}{g},\\
    A_{\mathrm{KV}}(T)
    &=\frac{H}{g}[a_{\alpha,B}(T+1)-1]
     =\frac{\alpha HT}{g}+\mathcal{O}\!\left(\frac{HB}{g}\right).
\end{aligned}
\end{equation}
All $g$ queries still use each selected entry, so
\begin{equation}
\begin{aligned}
    C_{\mathrm{attn}}
    &=\frac{H}{g}\,g\,\Phi_{\alpha,B}(T)
     =\frac{\alpha HT^2}{2}+\mathcal{O}(HBT),\\
    D_{\mathrm{meta}}(T)
    &=\frac{H}{g}\left\lfloor\frac{T}{B}\right\rfloor,\\
    C_{\mathrm{meta}}
    &=\frac{H}{g}[\Gamma_B(T)-T]
     =\frac{HT^2}{2gB}+\mathcal{O}\!\left(\frac{HT}{g}\right).
\end{aligned}
\label{eq:app-gqa-sa-work}
\end{equation}

\noindent\textbf{Shared indices versus shared scoring.}
Let $\chi_g$ denote the number of $d_h$-scale scoring operations used to form one group-block score, with $\chi_1=1$.
Aggregating separately computed scores from all $g$ queries gives $\chi_g=g$, as in the per-head score aggregation of NSA~\citep{Yuan2025NativeSA}.
A single dot product using a pooled group query instead gives $\chi_g=1$, but generally defines a different scoring rule.
Both produce one shared block set and have the same KV counts.
Their scoring arithmetic is $\Theta(\chi_gd_hD_{\mathrm{meta}})$ during decoding and $\Theta(\chi_gd_hC_{\mathrm{meta}})$ during prefill.
Shared block indices therefore do not by themselves imply shared scoring arithmetic.

\subsection{GQA+\method{}}
\label{app:complexity-gqa-namoh}

\noindent\textbf{Query-head routing.}
One option selects $K$ of the $H$ query heads independently.
Let $u_{t,j}=\mathbf 1\{\mathcal S_K(t)\cap\mathcal Q_j\neq\emptyset\}$ indicate whether token $t$ touches KV group $j$, where $\mathbf 1$ is the indicator function.
The number of KV groups written by a token satisfies
\begin{equation}
    \left\lceil\frac Kg\right\rceil
    \leq\sum_{j=0}^{H/g-1}u_{t,j}
    \leq\min\!\left(K,\frac Hg\right).
\end{equation}
If a shared pair is stored whenever any query in its group is selected, then $n_j=\sum_{t=0}^{T-1}u_{t,j}$ and
\begin{equation}
    M_{\mathrm{KV}}=\sum_jn_j,\qquad
    T\left\lceil\frac Kg\right\rceil
    \leq M_{\mathrm{KV}}
    \leq T\min\!\left(K,\frac Hg\right).
\end{equation}
Allowing active queries to read their groups' union histories gives $A_{\mathrm{KV}}(T)=\sum_j u_{T,j}n_j$.
Preserving separate query-head histories instead requires membership masks and makes access depend on overlaps between current and historical assignments.
Neither storage nor access is determined by $T$, $H$, $K$, and $g$ alone.

\noindent\textbf{Adopted KV-group routing.}
We instead select exactly $K/g$ of the $H/g$ KV groups.
Let $\mathcal G_t$ denote the selected group set for token $t$.
All $g$ query heads in each selected group participate and share one routing gate.
The group writes one KV pair.
Thus, exactly $K$ query and output projections and $K/g$ key and value projections are evaluated:
\begin{equation}
\begin{aligned}
    P_{\mathrm{total}}
    &=2H\left(1+\frac1g\right)d_md_h+\frac Hg d_m,\\
    P_{\mathrm{active}}
    &=2K\left(1+\frac1g\right)d_md_h+\frac Hg d_m.
\end{aligned}
\label{eq:app-group-parameters}
\end{equation}
For this routing rule, the group lengths satisfy
\begin{equation}
    n_j=\sum_{t=0}^{T-1}\mathbf 1\{j\in\mathcal G_t\},\qquad
    \sum_{j=0}^{H/g-1}n_j=\frac{TK}{g},\qquad
    \bar n=\frac{TK/g}{H/g}=\frac{TK}{H}.
\end{equation}
The group activation fraction remains $K/H$, so the average history length is $TK/H$, not $TK/(Hg)$.
Consequently,
\begin{equation}
    M_{\mathrm{KV}}=\frac{TK}{g},\qquad
    A_{\mathrm{KV}}(T)=\sum_{j\in\mathcal G_T}n_j
    \approx\frac{K}{g}\frac{TK}{H}
    =\frac{TK^2}{Hg}.
    \label{eq:app-group-kv}
\end{equation}

\noindent\textbf{Causal prefill derivation.}
For group loads $\boldsymbol n=(n_0,\ldots,n_{H/g-1})$, the squared coefficient of variation is
\begin{equation}
    \operatorname{CV}(\boldsymbol n)^2
    =\frac{(g/H)\sum_j(n_j-\bar n)^2}{\bar n^2}.
\end{equation}
Expanding the second moment gives
\begin{equation}
\begin{aligned}
    \sum_jn_j^2
    &=\frac Hg\bar n^2+\sum_j(n_j-\bar n)^2\\
    &=\frac{T^2K^2}{Hg}[1+\operatorname{CV}(\boldsymbol n)^2].
\end{aligned}
\label{eq:app-group-second-moment}
\end{equation}
Each local rank now produces $g$ queries, so
\begin{equation}
\begin{aligned}
    C_{\mathrm{attn}}
    &=\sum_j\sum_{r=1}^{n_j}gr
     =\frac g2\sum_jn_j^2+\frac g2\sum_jn_j\\
    &=\frac{T^2K^2}{2H}[1+\operatorname{CV}(\boldsymbol n)^2]
      +\frac{TK}{2}.
\end{aligned}
\label{eq:app-group-prefill}
\end{equation}
KV sharing therefore preserves \method{}'s balanced interaction count while reducing distinct KV storage and access by $g$.
When head-relative RoPE is enabled, all queries and the shared key use the same group-local position index.
Group routing also coarsens the routing choices and requires $g\leq K$; a single KV group leaves no nontrivial group selection.

\subsection{SA+\method{}}
\label{app:complexity-sa-namoh}

We first construct routed head histories, then form blocks in each head's local token order.
The retained fraction $\alpha$ applies to that local history, not to the original $T$-token prefix.
Parameter counts and KV storage remain those of \method{}.
Using the head lengths from Section~\ref{app:complexity-namoh},
\begin{equation}
\begin{aligned}
    A_{\mathrm{KV}}(T)
    &=\sum_{i\in\mathcal S_K(T)}[a_{\alpha,B}(n_i+1)-1]\\
    &=\alpha\sum_{i\in\mathcal S_K(T)}n_i+\mathcal{O}(KB)
      \approx\frac{\alpha TK^2}{H}.
\end{aligned}
\end{equation}
For prefill, apply the selected causal sum independently to every head:
\begin{equation}
\begin{aligned}
    C_{\mathrm{attn}}
    &=\sum_i\Phi_{\alpha,B}(n_i)\\
    &=\frac\alpha2\sum_i n_i(n_i+1)
      +\mathcal{O}\!\left(B\sum_i n_i\right)\\
    &=\frac{\alpha T^2K^2}{2H}[1+\operatorname{CV}(\boldsymbol n)^2]
      +\mathcal{O}(BTK).
\end{aligned}
\label{eq:app-sa-namoh-prefill}
\end{equation}
The indexer scans only the selected routed histories during decoding:
\begin{equation}
\begin{aligned}
    D_{\mathrm{meta}}(T)
    &=\sum_{i\in\mathcal S_K(T)}\left\lfloor\frac{n_i}{B}\right\rfloor
     =\frac1B\sum_{i\in\mathcal S_K(T)}n_i+\mathcal{O}(K),\\
    C_{\mathrm{meta}}
    &=\sum_i[\Gamma_B(n_i)-n_i]
     =\frac{1}{2B}\sum_i n_i^2+\mathcal{O}(TK)\\
    &=\frac{T^2K^2}{2HB}[1+\operatorname{CV}(\boldsymbol n)^2]
      +\mathcal{O}(TK).
\end{aligned}
\label{eq:app-sa-namoh-metadata}
\end{equation}
Under balance, the decoding scan has leading term $TK^2/(HB)$.
Routing reduces the histories scanned by the selector; block selection then reduces token-level reads within those histories.

\subsection{GQA+SA+\method{}}
\label{app:complexity-triple}

We combine KV-group routing with group-shared block selection.
Each token activates $K/g$ groups, and each active group selects blocks from its own routed history.
All $g$ queries in that group use the same selected blocks.
Parameter counts remain those in Equation~\ref{eq:app-group-parameters}, and
\begin{equation}
\begin{aligned}
    M_{\mathrm{KV}}&=\frac{TK}{g},\\
    A_{\mathrm{KV}}(T)
    &=\sum_{j\in\mathcal G_T}[a_{\alpha,B}(n_j+1)-1]\\
    &=\alpha\sum_{j\in\mathcal G_T}n_j
      +\mathcal{O}\!\left(\frac{KB}{g}\right)
      \approx\frac{\alpha TK^2}{Hg}.
\end{aligned}
\label{eq:app-triple-kv}
\end{equation}
Combining the local causal sum with the group second moment gives
\begin{equation}
\begin{aligned}
    C_{\mathrm{attn}}
    &=g\sum_j\Phi_{\alpha,B}(n_j)\\
    &=\frac{\alpha g}{2}\sum_jn_j(n_j+1)
      +\mathcal{O}\!\left(Bg\sum_jn_j\right)\\
    &=\frac{\alpha T^2K^2}{2H}[1+\operatorname{CV}(\boldsymbol n)^2]
      +\mathcal{O}(BTK).
\end{aligned}
\label{eq:app-triple-prefill}
\end{equation}
The shared indexer visits each candidate block once per active group:
\begin{equation}
\begin{aligned}
    D_{\mathrm{meta}}(T)
    &=\sum_{j\in\mathcal G_T}\left\lfloor\frac{n_j}{B}\right\rfloor
     =\frac1B\sum_{j\in\mathcal G_T}n_j
      +\mathcal{O}\!\left(\frac Kg\right),\\
    C_{\mathrm{meta}}
    &=\sum_j[\Gamma_B(n_j)-n_j]
     =\frac{1}{2B}\sum_jn_j^2
      +\mathcal{O}\!\left(\frac{TK}{g}\right)\\
    &=\frac{T^2K^2}{2HgB}[1+\operatorname{CV}(\boldsymbol n)^2]
      +\mathcal{O}\!\left(\frac{TK}{g}\right).
\end{aligned}
\label{eq:app-triple-metadata}
\end{equation}
Under balance, the decoding scan has leading term $TK^2/(HgB)$.
Relative to $H$-head MHA, persistent KV storage is $K/(Hg)$ and activation is approximately $\alpha K^2/(H^2g)$ of the baseline.
The factor $\alpha$ reduces activation, not persistent KV storage.

\subsection{Overall Costs and Scaling}
\label{app:complexity-overall}

\noindent\textbf{Persistent cache storage.}
For any SA variant, index its KV histories by $j=0,\ldots,H/g-1$, with $g=1$ for ungrouped attention.
The lengths $n_j$ equal $T$ without routing and are assignment-dependent otherwise.
Metadata contributes
\begin{equation}
    M_{\mathrm{meta}}
    =\nu d_h\sum_j\left\lceil\frac{n_j}{B}\right\rceil
    \leq\nu d_h\left(\frac{M_{\mathrm{KV}}}{B}+\frac Hg\right)
    \label{eq:app-general-metadata-storage}
\end{equation}
in scalars.
Total KV and metadata storage is $2d_hM_{\mathrm{KV}}+M_{\mathrm{meta}}$.
For the three-way combination, this is at most
\begin{equation}
    \frac{2d_hTK}{g}
    +\nu d_h\left(\frac{TK}{gB}+\frac Hg\right).
\end{equation}
Block allocation can additionally leave fewer than $B$ unused KV slots per nonempty history.
Metadata construction costs $\mathcal{O}(d_hM_{\mathrm{KV}})$ over the prefix and is dominated by projection work in this model.

\noindent\textbf{Arithmetic accounting.}
Let $h_{\mathrm{act}}$ be the number of active query heads: $H$ without head routing and $K$ for MoH and \method{} variants.
With group-shared access, the next token evaluates $gA_{\mathrm{KV}}(T)+h_{\mathrm{act}}$ query-key interactions, including self-attention.
Let $F_{\mathrm{prefill}}$ and $F_{\mathrm{decode}}$ denote total arithmetic work over prefill and one decoding step.
Assuming linear-work head and block selection,
\begin{equation}
\begin{aligned}
    F_{\mathrm{prefill}}
    &=\mathcal{O}\!\left(
        TP_{\mathrm{active}}
        +d_h[C_{\mathrm{attn}}+\chi_gC_{\mathrm{meta}}]
      \right),\\
    F_{\mathrm{decode}}
    &=\mathcal{O}\!\left(
        P_{\mathrm{active}}
        +d_h[gA_{\mathrm{KV}}(T)+h_{\mathrm{act}}
              +\chi_gD_{\mathrm{meta}}(T)]
      \right).
\end{aligned}
\label{eq:app-overall-work}
\end{equation}
Set $C_{\mathrm{meta}}=D_{\mathrm{meta}}=0$ without block selection.
For ungrouped attention, $\chi_1=1$.
Sorting all scores can add work beyond this linear-selection model.

\begin{table}[t]
    \centering
    \small
    \setlength{\tabcolsep}{7pt}
    \renewcommand{\arraystretch}{1.4}
    \caption{
        \textbf{Leading attention and indexer terms.}
        The prefill column reports $C_{\mathrm{attn}}+\chi_gC_{\mathrm{meta}}$;
        the decoding column reports $gA_{\mathrm{KV}}(T)+\chi_gD_{\mathrm{meta}}(T)$.
        Insert these terms into Equation~\ref{eq:app-overall-work} with the active weights in Table~\ref{tab:attention-complexity} to obtain total arithmetic costs.
        Routed histories are balanced, and finite-length and block-rounding terms are omitted here but retained in the derivations.
    }
    \label{tab:attention-arithmetic}
    \begin{tabular}{@{}lcc@{}}
        \toprule
        Mechanism & Prefill & Single-token decoding\\
        \midrule
        MHA
        & $HT^2/2$
        & $HT$\\
        GQA
        & $HT^2/2$
        & $HT$\\
        SA
        & $\left(\alpha+\frac1B\right)\frac{HT^2}{2}$
        & $\left(\alpha+\frac1B\right)HT$\\
        MoH
        & $KT^2/2$
        & $KT$\\
        \textbf{\method{}}
        & $T^2K^2/(2H)$
        & $TK^2/H$\\
        \midrule
        GQA+SA
        & $\left(\alpha+\frac{\chi_g}{gB}\right)\frac{HT^2}{2}$
        & $\left(\alpha+\frac{\chi_g}{gB}\right)HT$\\
        GQA+\textbf{\method{}}
        & $T^2K^2/(2H)$
        & $TK^2/H$\\
        SA+\textbf{\method{}}
        & $\left(\alpha+\frac1B\right)\frac{T^2K^2}{2H}$
        & $\left(\alpha+\frac1B\right)\frac{TK^2}{H}$\\
        GQA+SA+\textbf{\method{}}
        & $\left(\alpha+\frac{\chi_g}{gB}\right)\frac{T^2K^2}{2H}$
        & $\left(\alpha+\frac{\chi_g}{gB}\right)\frac{TK^2}{H}$\\
        \bottomrule
    \end{tabular}
\end{table}

For the three-way combination with balanced histories and $\alpha TK/H\gg B$, Equation~\ref{eq:app-overall-work} becomes
\begin{equation}
\begin{aligned}
    F_{\mathrm{prefill}}
    &=\mathcal{O}\!\left(
        T\left[2K\left(1+\frac1g\right)d_md_h+\frac Hg d_m\right]
        +\frac{d_hT^2K^2}{H}
         \left[\alpha+\frac{\chi_g}{gB}\right]
      \right),\\
    F_{\mathrm{decode}}
    &=\mathcal{O}\!\left(
        2K\left(1+\frac1g\right)d_md_h+\frac Hg d_m
        +\frac{d_hTK^2}{H}
         \left[\alpha+\frac{\chi_g}{gB}\right]
      \right).
\end{aligned}
\label{eq:app-triple-total-work}
\end{equation}
With per-head score aggregation, $\chi_g=g$, so KV sharing does not reduce the leading indexer arithmetic by $g$.
A pooled-query indexer with $\chi_g=1$ realizes that additional scoring reduction.
Neither case changes the $g$ separate attention computations per shared KV entry.

\noindent\textbf{Fractional versus fixed budgets.}
The multiplicative activation formulas assume a fixed retained fraction of each local history.
They do not imply the same gain under a fixed absolute budget.
If at most $R$ blocks are retained per active history and $S=RB$ is the corresponding entry budget, the group-routed hybrid instead satisfies
\begin{equation}
    A_{\mathrm{KV}}(T)\leq\frac{K}{g}S,\qquad
    C_{\mathrm{attn}}\leq TKS.
\end{equation}
Once histories exceed this budget, head routing does not multiply these saturated bounds by another factor of $K/H$.
It still reduces storage and the histories scanned by the indexer.

At fixed $\alpha$, $B$, $H$, $K$, and $g$, both fractional attention and full metadata scanning remain quadratic in prefill length $T$.
Their coefficients are reduced, not their asymptotic order.
Setting $\alpha=1$ restores dense attention within each available local history, and block selection can then be bypassed.
These costs describe computational structure rather than guaranteed latency; physical gains also depend on cache reuse, data movement, and kernel scheduling.

\end{document}